\documentclass[10pt]{article}
\PassOptionsToPackage{dvipsnames,svgnames,x11names}{xcolor}
\usepackage[citestyle=authoryear]{colab}

\usepackage{longtable}
\usepackage{calc}
\usepackage{array}
\usepackage{multirow}
\usepackage{graphicx}
\usepackage{float}
\usepackage{algorithm}
\usepackage{algorithmic}
\graphicspath{{figures/}}

\definecolor{rtpTableHighlight}{gray}{0.93}
\newcolumntype{L}{>{\columncolor{white}[0pt][\tabcolsep]}l}
\newcolumntype{C}{>{\columncolor{white}[\tabcolsep][0pt]}c}
\usepackage{amsmath,amsfonts,bm}

\def\eqref#1{equation~\ref{#1}}

\def\1{\bm{1}}

\DeclareMathAlphabet{\mathsfit}{\encodingdefault}{\sfdefault}{m}{sl}
\SetMathAlphabet{\mathsfit}{bold}{\encodingdefault}{\sfdefault}{bx}{n}

\newcommand{\RmbNetPoints}{4.889}

\newcommand{\DefaultFalseFresh}{814}
\newcommand{\DefaultAcceptable}{5618}

\newcommand{\RoutingGain}{1.25}
\newcommand{\BinaryGain}{2.25}

\newcommand{\RecurrentMatchedGain}{0.125}
\newcommand{\RecurrentMatchedLower}{-2.50}
\newcommand{\RecurrentMatchedUpper}{2.75}

\newcommand{\RtpMeanSteps}{7.78}
\newcommand{\FixedTenMeanSteps}{12.33}
\newcommand{\RtpMeanCallSeconds}{1.051}
\newcommand{\FixedTenMeanCallSeconds}{1.115}

\newcommand{\RtpPninetyFiveSeconds}{1.294}
\newcommand{\FixedTenPninetyFiveSeconds}{1.268}

\newcommand{\NominalSourceGain}{2.8}

\newcommand{\HoldEightSourceGain}{2.9}

\newcommand{\RtpStepReduction}{36.9}
\newcommand{\MainRtpRate}{48.6}
\newcommand{\MainRmbRate}{84.8}
\newcommand{\MainNetGain}{8.2}
\newcommand{\MainRmbNetGain}{4.9}
\newcommand{\MainBridgeGain}{5.8}
\newcommand{\MainExtraLearningGain}{3.1}
\newcommand{\MainMatchedGain}{5.1}
\newcommand{\CurrentPrefixGain}{2.5}
\newcommand{\CurrentPrefixLower}{0.0}
\newcommand{\CurrentPrefixUpper}{5.0}

\newcommand{\RoboTauV}{0.1270}
\newcommand{\RoboTauA}{0.1960}
\newcommand{\RoboBeta}{2.1291}
\newcommand{\RmbTauV}{0.1068}
\newcommand{\RmbTauA}{0.1604}
\newcommand{\RmbBeta}{2.1114}
\newcommand{\RecurrentExceedanceGain}{-1.04}
\newcommand{\RecurrentExceedanceLower}{-1.46}
\newcommand{\RecurrentExceedanceUpper}{-0.64}
\newcommand{\RoboDeltaWidth}{3272}
\newcommand{\RoboStateBytes}{589,824}
\newcommand{\RoboBridgeFiveBytes}{393,216}
\newcommand{\RoboStateMiB}{0.5625}
\newcommand{\RmbDeltaWidth}{3520}
\newcommand{\RmbStateBytes}{552,960}
\newcommand{\RmbBridgeFiveBytes}{368,640}
\newcommand{\RmbStateMiB}{0.52734375}
\newcommand{\SelectorWidth}{3219}
\newcommand{\ResidualParameters}{7,374,528}
\newcommand{\SelectorParameters}{20,743,240}
\newcommand{\RoboBridgeParameters}{17,126,592}
\newcommand{\RoboTotalParameters}{37,869,832}
\newcommand{\RmbBridgeParameters}{17,507,520}
\newcommand{\RmbTotalParameters}{38,250,760}
\newcommand{\BridgeMatchedGain}{3.875}
\newcommand{\BridgeMatchedLower}{1.125}
\newcommand{\BridgeMatchedUpper}{6.500}

\providecommand{\tightlist}{\setlength{\itemsep}{0pt}\setlength{\parskip}{0pt}}

\title{Revision, Not Restart: Revisable Visual Plans\\
for Closed-Loop World--Action Models}
\author{Pengyiang Liu$^{1,*}$ \quad Junbo Niu$^{2,*}$ \quad Wenhao Zheng$^{1}$ \quad Xinchen Chen$^{1}$\\
Canyu Li$^{1}$ \quad Zhongyue Shi$^{1}$ \quad Jiahao Xie$^{1}$ \quad Si Liu$^{1}$\\[0.2em]
\textsuperscript{1}Beihang University \quad \textsuperscript{2}Peking University\\
\textsuperscript{*}Equal contribution}
\labname{Colab}
\institution{}
\colabdate{2026-09-28}
\paperurl{https://buaa-colalab.github.io/RTP/}
\githuburl{}
\huggingfaceurl{}
\dataurl{}
\colabrunningtitle{RTP: Revisable Temporal Planning}
\hypersetup{pdftitle={Revision, Not Restart: Revisable Visual Plans for Closed-Loop World--Action Models}}

\begin{document}
\maketitle
\vspace{-10pt}
\makeatletter
\begin{colababstract}
World--action models use predicted visual futures to condition robot actions, yet execution feedback can invalidate parts of a prediction while leaving its task structure useful. We propose \textbf{R}evisable \textbf{T}emporal \textbf{P}lanning (RTP), which maintains the visual future as a persistent action condition and revises it after feedback. Its central mechanism is a learned revision bridge: it resumes an intermediate state saved during visual generation and adapts its continuation to current observations. Visual and action supervision connect this revision to subsequent control. Time-aware history supplies observed evidence, and an adaptive policy selects retention, bridge revision, or fresh replanning from new noise before decoding the next action. On RoboMME and RMBench, RTP achieves task-averaged success rates of \MainRtpRate\% and \MainRmbRate\%, respectively. Matched comparisons support learned continuation; estimated checkpoint-source and action-prefix effects are positive but less precisely resolved. These results connect feedback-driven visual-plan revision to closed-loop task performance.

\vspace{0.6em}
{\small\textbf{Project Page:}~\url{\@paperurl}}
\end{colababstract}
\makeatother
\vspace{0.5em}

\begin{center}
\begin{minipage}{\textwidth}
\centering
\includegraphics[width=0.94\textwidth]{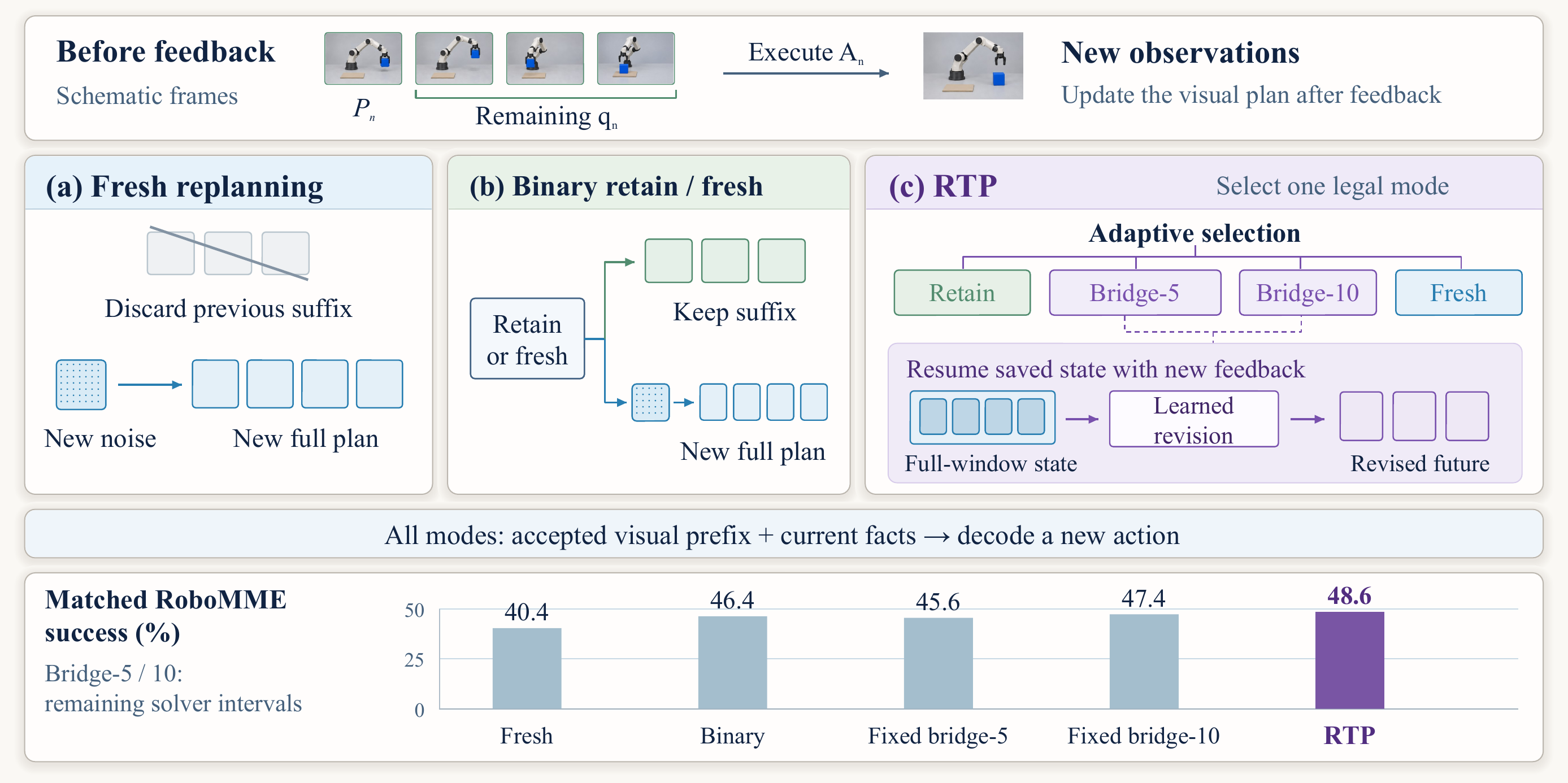}
\captionof{figure}{\textbf{Visual-plan updates.} \emph{Top:} predicted grasp versus feedback with the cube still on the table. \emph{Middle:} fresh replanning, binary reuse/fresh selection, and RTP revision. \emph{Bottom:} matched RoboMME success. All modes re-decode actions; fixed-depth policies use fresh when required states are unavailable.}
\label{fig:rtp-cycle}
\end{minipage}
\end{center}

\section{Introduction}
\label{sec:intro}
Vision--language--action (VLA) policies transfer visual and language representations to robot control \citep{openvla,pi05}. World--action models (WAMs) additionally learn visual dynamics alongside actions, drawing on video-based priors for physical interaction \citep{lingbotva,dreamzero}. In models that decode actions from predicted visual states, the generated future serves as a plan condition: it helps determine the robot's next action. Maintaining this condition after execution is therefore part of the control problem.

Execution introduces observations that can disagree with this predicted future. A contact may occur later than expected even when the planned approach and subsequent manipulation remain useful. Reusing the prediction unchanged can preserve the timing error, whereas generating a new future from noise repeats the visual-generation process. The remaining prediction offers a starting point for feedback-driven updating, provided its continuation can be reconciled with the new observation.

Recent work addresses complementary aspects of this problem. Persistent memory retains earlier observations \citep{memorywam}, while efficient WAMs reduce the cost of obtaining representations for action generation \citep{fastwam,rift}. Feedback-aware execution adjusts how long a plan is followed \citep{ffdc} or adapts reused planner context to current observations \citep{ahawam}. Diffusion planners further show how existing trajectories can be repaired or continued across observations \citep{zhou2023adaptive,hoeg2024streaming}. These developments motivate a central question: \textbf{how can a WAM revise an already generated visual future under execution feedback and use it to guide the next action?}

Our key insight is to maintain the visual future as a persistent action condition that evolves with execution feedback. Saved intermediate denoising states provide starting points within the generation process that produced the plan. These states were formed under earlier observations; continuing them under current feedback requires adapting the remaining generation. We learn this transition with visual and action supervision so that the revised future can guide the next action.

Revisable Temporal Planning (RTP) implements this feedback-driven update through a learned revision bridge. Time-aware history supplies recent detail and older evidence at their original environment timestamps, while a separate record preserves the visual plan and its intermediate generation states. An adaptive policy allocates revision effort by selecting retention, bridge revision, or fresh replanning from newly sampled noise, using estimated differences from fresh generation. The accepted visual plan and current observations then condition the next action. We instantiate RTP on task-adapted LingBot-VA~\citep{lingbotva} and evaluate it on RoboMME~\citep{robomme} and RMBench~\citep{rmbench}.

Our contributions are:
\begin{itemize}
\tightlist
\item A persistent visual action condition that is revised across feedback alongside a separate, time-aware record of observed history.
\item A learned revision bridge that adapts saved visual-generation states to current feedback through visual and action supervision.
\item Closed-loop comparisons separating learned correction, checkpoint source, and action conditioning, together with success--cost comparisons of complete update policies.
\end{itemize}

\section{Related Work}
\subsection{World--Action Models and Future Conditioning}
Diffusion transformers and flow matching support scalable visual and action generation \citep{peebles2022dit,lipman2022flowmatching}. Joint models couple future observations with robot actions \citep{zhu2025uwm,lingbotva,dreamzero,motus}, with native causal pretraining extending this interface \citep{lingbotva2}. Predicted action conditions include latent subgoals and addressable object states \citep{lawam,oawam}; visual action representations include multiview action images and optical flow \citep{actionimages,flowwam}. RGB-D prediction \citep{xwam} and action-conditioned self-motion supervision \citep{selfwam} further structure the visual future.

Efficient models reduce rollout, action-decoding, and future-conditioning costs \citep{fastwam,lightwam,fasterwamdot,fasterwamfuture,rift}. Asynchronous systems align predicted context with action execution \citep{ahawam,futurertc}. These methods improve generation or execution efficiency; RTP addresses how the visual condition itself changes after feedback, continuing saved solver states and decoding actions from the accepted clean prefix.

\subsection{Feedback Verification and Generative-Plan Revision}
Diffusion policies generate action sequences with receding-horizon execution \citep{diffusionpolicy}. Revising a generated sample can start from a re-noised endpoint \citep{meng2021sdedit}; token-wise noise levels and selective re-noising support incremental and cross-horizon generation \citep{chen2024diffusionforcing,kim2026diffusionreroll}. Adaptive Online Replanning uses plan likelihood to choose retention, repair through re-noising, or resampling \citep{zhou2023adaptive}, while Streaming Diffusion Policy recursively updates a partially denoised action buffer under new observations \citep{hoeg2024streaming}.

Prediction--observation consistency supports adaptive execution and online world-model correction \citep{ffdc,feedbackworldmodel}. Action-conditioned rollout search refines proposed plans \citep{worldactionplanner}. RTP resumes intermediate states saved while generating the existing visual plan, learns their continuation under new observations, and uses the revised visual prefix to condition the next action.

\subsection{Long-Horizon World Memory}
Long-horizon video understanding and generation require memory of evolving scenes. Streaming counting and spatial/online understanding probe state maintenance across observations \citep{svcbench,ovosbench,ovobench}, while long-video evidence retrieval connects answers to earlier events \citep{trace}. Video world models preserve history through consolidation, addressable caches, and retrieval \citep{fademem,worldtrace,worldkv}.

Robotic memory combines perceptual features and semantic summaries \citep{memoryvla,mem}, retrieves relevant experience \citep{memer,chameleon}, preserves spatial context \citep{sam2act}, and tracks progress in cyclic tasks \citep{cyclemanip}. Persistent world--action memory supports action generation from long histories \citep{memorywam}. RTP maintains both observed history and a predicted future: multiscale, environment-time-addressed records supply evidence, while a separate predictive record is revised after feedback to condition control.

\section{Method}
\label{sec:method}
\begin{figure}[t]
\centering\includegraphics[width=\linewidth]{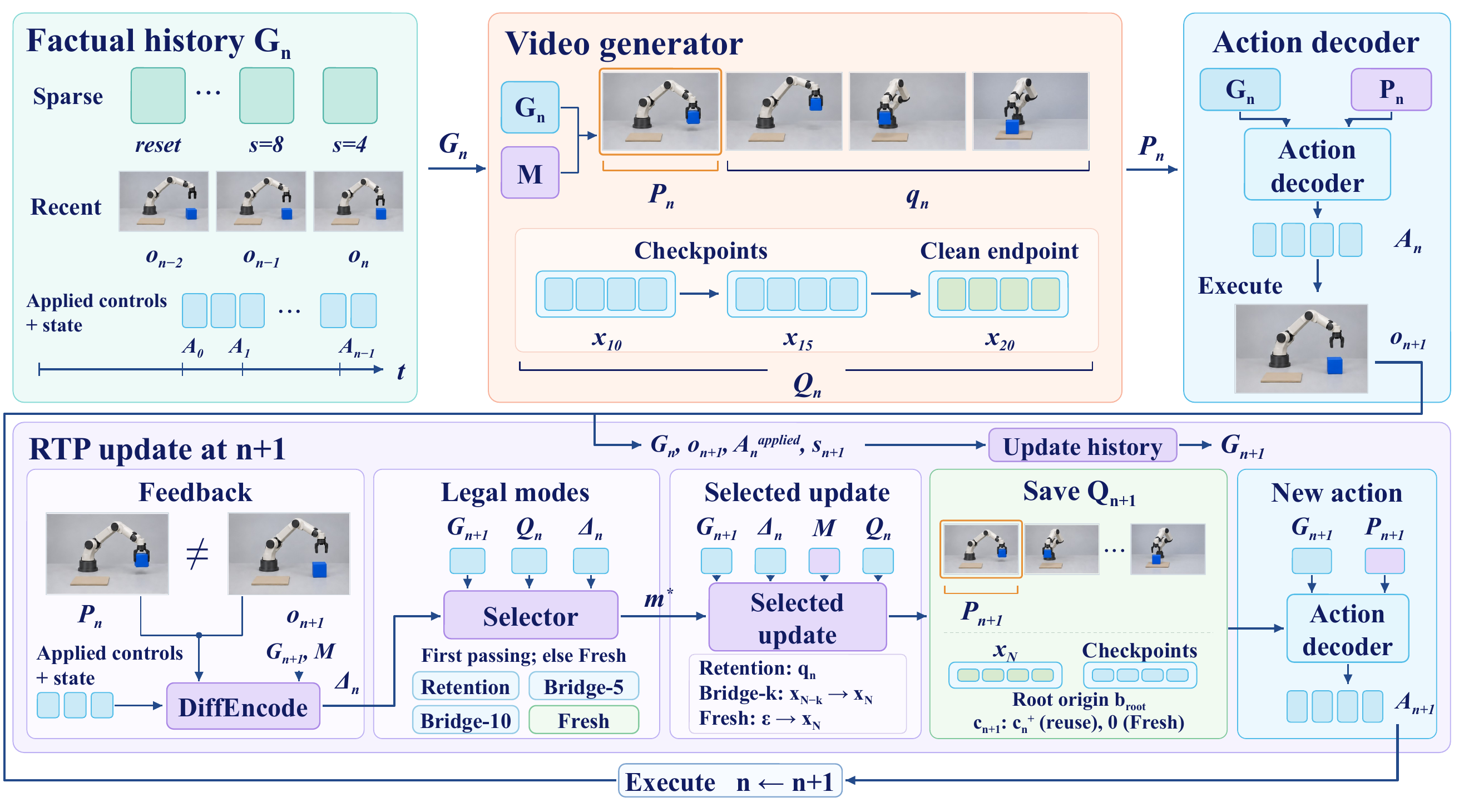}
\caption{\textbf{The RTP feedback loop.} The upper row illustrates a fresh plan root. At each feedback boundary, update factual history, compare prediction with feedback, and run one selected visual update. Save the accepted record and decode a new action from its next prefix and current facts.}
\label{fig:rtp-framework}
\end{figure}

RTP maintains observed history and an unexecuted visual plan. At each feedback boundary, it updates the history, selects a plan update, and decodes a new action under current observations (Figure~\ref{fig:rtp-framework}). Training first adapts LingBot-VA~\citep{lingbotva} to time-aware history (Stage A), then learns feedback-conditioned revision with the adapted model frozen (Stage B), and finally fits a discrepancy estimator with the preceding modules fixed (Stage C).

\suppressfloats[t]

\subsection{Time-Aware History and Persistent Plans}
\label{sec:foundation}
\textbf{Factual history.}
A control boundary $b_n$ is the physical sample time when feedback arrives from an executed action block; initialization uses the reset observation. One temporal group spans $J$ native environment samples. Its factual record contains all-camera causal latents $z_i$, applied controls $A_{i-1}^{\rm applied}$ leading to that endpoint, proprioception $s_i$, and sample index $\rho_i$:
\[
r_i=(z_i,A_{i-1}^{\rm applied},s_i,\rho_i).
\]
The causal encoder processes the continuous observation stream before history selection; each selected group retains its original timestamp.

The \emph{temporal pyramid} preserves recent detail and older evidence within a fixed group budget. For lookback span $q_\ell$ and sampling stride $s_\ell$, both measured in groups, with phase $i_0$ fixed at reset, its candidates are
\begin{equation}
\mathcal K_\ell(n)=\left\{i\le n:0\le n-i<q_\ell,\ (i-i_0)\bmod s_\ell=0\right\}.
\label{eq:pyramid-main}
\end{equation}
The sampler reserves the initial observation as a reset anchor and a recent-group quota, adds unused pyramid candidates newest first, and fills spare capacity with the newest omitted groups. Records are ordered by physical endpoint. The primary spans/strides are $(12,1),(76,4),(204,8)$ (Appendix~\ref{app:foundation}).

Let $M$ denote fixed instruction/reference context. The selected record indices $\mathcal I_n$ determine the input sequence and its derived layer-wise keys and values:
\begin{equation}
\begin{aligned}
U_n&=\operatorname{Pack}\left(M,\{r_i\}_{i\in\mathcal I_n};\Pi_n,\mathcal M_n\right),\\
\mathcal C_n=\{K_n^{(\ell)},V_n^{(\ell)}\}_{\ell=1}^{L}
&=\operatorname{Prefill}_{\psi}\left(U_n;\Pi_n,\mathcal M_n\right).
\end{aligned}
\label{eq:ground}
\end{equation}
Pack organizes context at positions $\Pi_n$ with visibility mask $\mathcal M_n$. Prefill, with adapted parameters $\psi$, computes attention keys $K_n^{(\ell)}$ and values $V_n^{(\ell)}$ for $L$ backbone layers. The factual interface $G_n=(U_n,\Pi_n,\mathcal M_n,\mathcal C_n)$ combines these inputs and their KV conditioning; explicit $M$ arguments identify the same packed context. Changes in facts or coordinate views require rebuilding the corresponding KV.

\textbf{Predictive state.}
\emph{Fresh replanning} generates a new $H$-group visual plan from independent noise, conditioned on $G_n$ and $M$. It establishes a \emph{plan root}: the fixed prediction window and its generation provenance. One action call consumes $H_c$ groups. Generation spans $N=20$ noise-to-clean Euler intervals \citep{lipman2022flowmatching}; $x_j$ denotes the full window at solver index $j$ and $x_N$ its clean endpoint. Its unconsumed portion is
\begin{equation}
\hat z_n=(P_n,q_n),\qquad
P_n=\operatorname{Prefix}_{H_c}(\hat z_n),\qquad L_n=H-c_n.
\label{eq:state}
\end{equation}
Here $c_n$ counts consumed groups, $L_n$ is the remaining length, and Prefix extracts $H_c$ groups for the next action; $q_n$ contains the rest. The persistent record $Q_n$ stores $x_N$, checkpoints $\mathcal S_n$, root origin $b_{\rm root}$, consumed index, and provenance (Appendix~\ref{app:state}). With $H=4,H_c=1$, one executed group leaves three active groups; checkpoints retain the full four-group window. Execution advances the count to $c_n^+=c_n+H_c$ and contributes new facts to $G_{n+1}$. Reuse requires $H-c_n^+\ge H_c$.

\textbf{Time coordinates.}
Factual records use current-boundary positions $p_i^{\rm fact}=(\rho_i-b_n)/J$, while a saved plan retains root-relative positions $p_t^{\rm plan}=(\tau_t-b_{\rm root})/J$, where $\tau_t$ is a predicted group's physical sample endpoint. Restoration rebuilds the fact/query coordinate view for that root. Action decoding expresses the selected prefix relative to the current boundary, preserving the stored solver root.

\textbf{Stage A: model adaptation to time-aware history.}
We fine-tune LingBot-VA's visual and action generation modules on demonstration futures and aligned actions conditioned on the selected history, keeping observation and language encoders fixed. The action branch uses observed future prefixes during training and generated prefixes at deployment. The adapted model and history conventions form the \emph{temporal foundation}; objectives and sampling mixtures appear in Appendix~\ref{app:training}.

\subsection{Feedback-Conditioned Plan Revision}
\label{sec:revision}
The \emph{feedback-conditioned revision bridge} resumes a saved generation state under new observations. A learned correction augments the frozen visual velocity field to revise the unexecuted plan. At $b_{n+1}$, its feedback descriptor aligns the executed prefix with the new observation:
\begin{equation}
\begin{aligned}
\Delta_n=\operatorname{DiffEncode}\bigl(&\operatorname{Endpoint}_{b_{n+1}}(P_n),E_v(o_{n+1}),A_n^{\rm applied},\\
&s_n,s_{n+1},\operatorname{GroundSummary}(G_{n+1}),M\bigr).
\end{aligned}
\label{eq:feedback}
\end{equation}
Endpoint selects the prediction at $b_{n+1}$; $E_v$ encodes observation $o_{n+1}$, and GroundSummary pools factual conditioning. DiffEncode is a multilayer perceptron (MLP) combining pooled predicted/observed latents and their difference, applied controls, old/new proprioception and its change, and factual/task summaries. Its output $\Delta_n$ conditions revision and selection.

\textbf{Resuming a saved state.}
A checkpoint $S_{n,j}$ stores $x_j$, solver time, and root/source metadata before interval $j$. Bridge-$k$ restores $S_{n,N-k}$ and executes the final $k\in\{5,10\}$ intervals:
\begin{equation}
\hat z_{n+1}^{(k)}
=B_\eta^{(k)}(S_{n,N-k},G_{n+1},\Delta_n,M).
\label{eq:bridge}
\end{equation}
With $N=20$, Bridge-5 resumes $x_{15}$ for five intervals; Bridge-10 resumes $x_{10}$ for ten. Both share one correction network with a zero-initialized output layer. The post-execution feature \mbox{$a_n=c_n^+/H$} locates the active range. Under current facts,
\begin{samepage}
\begin{equation}
x_{j+1}=x_j+h_j\left[
v_\theta(x_j,u_j,G_{n+1},M)
+r_\eta(h_\theta(x_j,u_j,G_{n+1},M),\Delta_n,u_j,a_n)\right].
\label{eq:residual-bridge}
\end{equation}
Here $u_j$ is solver time, $h_j=u_{j+1}-u_j$ its step size, $v_\theta$ the frozen visual velocity field, $h_\theta$ its final visual representation, and $r_\eta$ the learned velocity correction. Restoration evolves the complete saved window under recomputed attention, then returns its unexecuted timestamps. Consumed positions remain solver variables; observed evidence enters through $G_{n+1}$ (Appendix~\ref{app:state}).
\end{samepage}

\textbf{Stage B: revision-bridge training.}
Stage B trains the feedback encoder DiffEncode and velocity correction $r_\eta$ with the adapted model fixed. A fresh controller replans at every boundary and archives each generated root. Archived roots are paired with later feedback after $c=1,2,3$ consumed groups, leaving $3,2,1$ groups. Successful and failed behavior trajectories supply observed continuations $Z^*_{n+1}$ and next continuous commands $A^*_{n+1}$ before actuator conversion.

Both bridge depths and an independently generated fresh reference $\hat z_{n+1}^{\mathrm{fresh}}$ use the same current facts. Their candidate prefixes are $P_{n+1}^{(k)}=\operatorname{Prefix}_{H_c}(\hat z_{n+1}^{(k)})$. The frozen action decoder $\pi_\phi$ produces $A_{n+1}^{(k)}=\pi_\phi(G_{n+1},P_{n+1}^{(k)},M)$. Visual and action distances $d_v,d_a$ are variance-scaled mean squared errors, with continuous denormalized action coordinates; $\operatorname{sg}$ stops gradients. We fit observed continuation and behavior actions, with fresh consistency regularization:
\begin{samepage}
\begin{equation}
\begin{aligned}
\mathcal L_B^{(k)}
&=d_v(\hat z_{n+1}^{(k)},\operatorname{sg}(Z^*_{n+1});I_*)
+d_a(A_{n+1}^{(k)},\operatorname{sg}(A^*_{n+1}))\\
&\quad+0.1d_v(\hat z_{n+1}^{(k)},\operatorname{sg}(\hat z_{n+1}^{\mathrm{fresh}});I_T),
\qquad
\mathcal L_B=\tfrac12(\mathcal L_B^{(5)}+\mathcal L_B^{(10)}).
\end{aligned}
\label{eq:bridge-loss}
\end{equation}
$I_*$ and $I_T$ are shared physical timestamps with the observed continuation and fresh reference, respectively. Archived states and targets are detached. Gradients through the frozen visual solver and action decoder train only the residual and DiffEncode (Appendix~\ref{app:distances}).
\end{samepage}

\subsection{Adaptive Updates and Action Generation}
\label{sec:selection}
The update selector chooses retention, Bridge-5, Bridge-10, or fresh replanning, denoted $\{0,5,10,T\}$. \emph{Retention} reuses the remaining visual prediction with zero visual solver updates; bridge modes revise it as in Section~\ref{sec:revision}. All modes re-decode actions under current facts. Legal reuse modes $\mathcal V_n\subseteq\{0,5,10\}$ require at least $H_c$ remaining groups and, for revision, the corresponding checkpoint.

\textbf{Stage C: discrepancy-estimator fitting and calibration.}
With the adapted model and bridge frozen, all legal candidates and two independent fresh references $T_1,T_2$ are expanded offline under common facts and an action-noise seed. For visual/action modality $r\in\{v,a\}$, distances $y_r^m=d_r(m,T_1)$ compare mode $m$'s output with fresh reference $T_1$ and supervise estimated discrepancies $\widehat d_r^m\ge0$ and estimation-error scales $\widehat u_r^m>0$. The estimator reads $\Delta_n$, factual summaries, and saved-state statistics. Disjoint calibration data set tolerances $\tau_r$ at the fresh--fresh 90th percentiles and the nonnegative multiplier $\beta$ at the standardized-error 95th percentile (Appendix~\ref{app:distances}).

Default fitting and calibration use archived fresh-parent states: saved fresh plans paired with later feedback. Appendix~\ref{app:repeats} evaluates recurrent deployment states reached after RTP's accepted updates, and recurrent fitting of the same correction and estimator on tuples collected from those states.

\textbf{Online selection and execution.}
Before visual generation, the estimator scores legal modes. The first passing mode in the priority order retention, Bridge-5, Bridge-10 is selected:
\begin{equation}
\begin{aligned}
\mathcal F_n&=\{m\in\mathcal V_n:
\widehat d_r^m+\beta\widehat u_r^m\le\tau_r
\quad\forall r\in\{v,a\}\},\\
m^*&=\begin{cases}
\min_{0\prec5\prec10}\mathcal F_n,&\mathcal F_n\ne\varnothing,\\
T,&\mathcal F_n=\varnothing .
\end{cases}
\end{aligned}
\label{eq:route}
\end{equation}
The chosen branch produces $\hat z_{n+1}$; an empty passing set triggers fresh. Every mode extracts $P_{n+1}=\operatorname{Prefix}_{H_c}(\hat z_{n+1})$ and decodes a new action:
\begin{equation}
A_{n+1}=\pi_\phi(G_{n+1},P_{n+1},M).
\label{eq:action}
\end{equation}
Fresh establishes a new root with $c_{n+1}=0$; reuse keeps the root with $c_{n+1}=c_n^+$. A bridge saves the corrected window, immutable starting checkpoint, and newly visited required checkpoints; retention preserves the existing set. Appendix~\ref{app:state} specifies checkpoint availability across updates.

\begin{algorithm}[!t]
\caption{Recurrent RTP update with fixed model parameters}\label{alg:rtp}
\begin{algorithmic}[1]
\STATE Initialize $n=0$, facts $G_0$, a fresh root $Q_0$, and action $A_0$.
\LOOP
\STATE Execute $A_n$; stop if the episode terminates.
\STATE Append observed/applied records once and rebuild $G_{n+1}$.
\STATE Compute feedback $\Delta_n$ (Eq.~\ref{eq:feedback}); advance $c_n^+=c_n+H_c$.
\STATE Construct legal modes $\mathcal V_n$; select $m^*$ using Eq.~\ref{eq:route}.
\STATE Rebuild positions/masks and run the selected visual update.
\STATE Save the accepted record $Q_{n+1}$ and its checkpoints.
\STATE Extract $P_{n+1}$; decode $A_{n+1}$ (Eq.~\ref{eq:action}); $n\gets n+1$.
\ENDLOOP
\end{algorithmic}
\end{algorithm}

\section{Experiments}
\label{sec:experiments}
\subsection{Experimental Setup}
\textbf{Benchmarks and training.}
RoboMME has 16 tasks across Counting, Permanence, Reference, and Imitation; RMBench has nine dual-arm tasks \citep{robomme,rmbench,robotwin}. Evaluation uses 50/100 resets per task (800/900 total), $H=4,H_c=1$, and action lengths $J=4/16$. Within each run, matched update policies share adapted weights, history, normalization, and action interfaces. History controls re-adapt their bases and corrections at equal data/update budgets (Appendix~\ref{app:training}).

\textbf{Baselines.}
Baselines cover vision--language--action and world--action models (Tables~\ref{tab:main-robomme} and~\ref{tab:main-rmbench}); benchmark configurations are detailed in Appendix~\ref{app:controls}. MME-VLA~\citep{robomme} uses frame sampling (FrameSamp) or test-time training (TTT), both with modulator integration. The dense-history and time-aware LingBot-VA variants use fresh replanning. Published benchmark entries retain their reported configurations; the within-model controls provide matched comparisons.

\textbf{Evaluation.}
Evaluation includes all resets with equal task weights. A reset key identifies an episode's initial environment state and randomization. Paired 95\% intervals resample these keys 10,000 times within task, clustering repeated rollout seeds. Fitting, calibration, cost matching, diagnostics, and test use disjoint trajectory pools. Policies share reset/initial randomness and follow their own later feedback. Paired contrasts use unrounded counts. Reported intervals characterize reset variation conditional on the fitted model, rather than variation across independent training runs.

\begin{figure}[!t]
\centering\includegraphics[width=\linewidth]{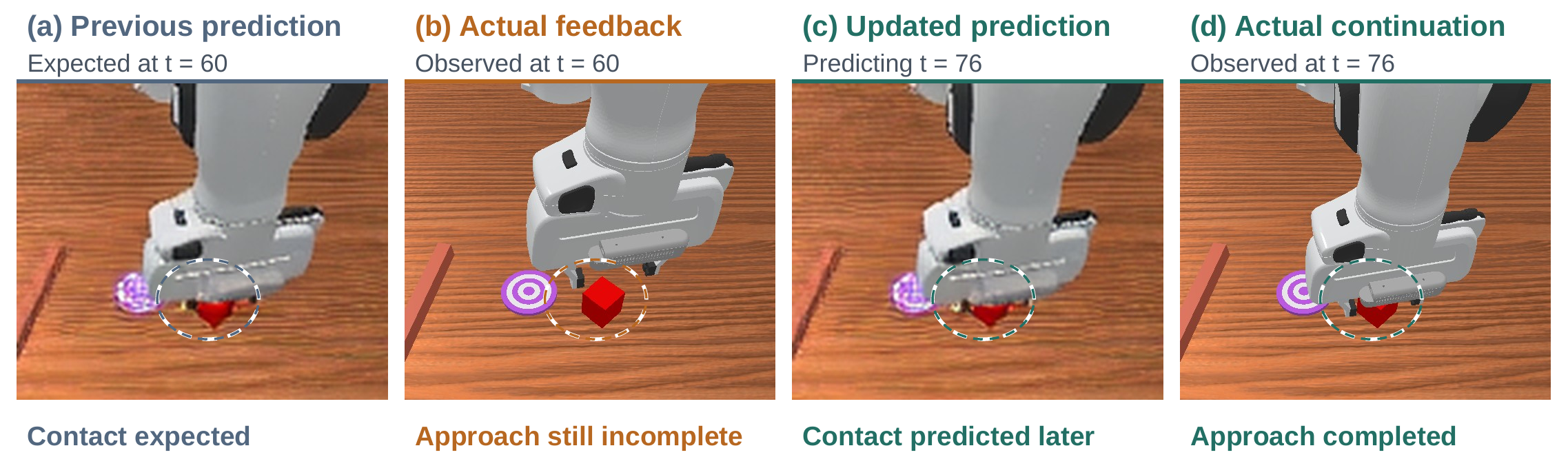}
\caption{\textbf{Prediction and execution.} Base-model continuation under a joint-target hold at samples 52--59. Prediction (a) and execution observation (b) disagree at 60; the updated prediction for 76 (c) better matches the approach progress in the subsequent observation (d).}
\label{fig:delayed-contact}
\end{figure}

\subsection{Main Results}
RTP achieves the highest observed task-averaged success among the compared methods on both RoboMME and RMBench (Tables~\ref{tab:main-robomme} and~\ref{tab:main-rmbench}). With the same time-aware model, updating persistent visual plans improves success over fresh replanning by \MainNetGain{} and \MainRmbNetGain{} percentage points, respectively, based on the displayed rates. Task-level comparisons include a tie and a regression on RoboMME, and several intervals include zero (Appendix~\ref{app:controls}). The following controls examine learned correction, the starting state, and the visual input to action decoding.

\begin{table}[!htbp]\centering\small
\caption{Task-averaged success on RoboMME.}\label{tab:main-robomme}
\begin{tabular}{@{}LC@{}}\toprule
Method & Success (\%) $\uparrow$\\\midrule
\multicolumn{2}{@{}l}{\emph{Vision--language--action models}}\\
$\pi_{0.5}$ \citep{pi05,robomme} & 17.9\\
MME-VLA (TTT--Modul) \citep{robomme} & 22.0\\
MemER \citep{memer,robomme} & 42.4\\
MME-VLA (FrameSamp--Modul) \citep{robomme} & 44.5\\
\midrule
\multicolumn{2}{@{}l}{\emph{World--action models}}\\
Fast-WAM \citep{fastwam} & 11.3\\
LingBot-VA (dense history) & 33.5\\
LingBot-VA (time-aware) & 40.4\\
\rowcolor{rtpTableHighlight}
\textbf{RTP (Ours)} & \textbf{48.6}\\
\bottomrule\end{tabular}
\end{table}

\begin{table}[!htbp]\centering\small
\caption{Task-averaged success on RMBench.}\label{tab:main-rmbench}
\begin{tabular}{@{}LC@{}}\toprule
Method & Success (\%) $\uparrow$\\\midrule
\multicolumn{2}{@{}l}{\emph{Vision--language--action models}}\\
$\pi_{0.5}$ \citep{pi05,rmbench} & 10.4\\
X-VLA \citep{xvla,rmbench} & 9.8\\
Mem-0 \citep{rmbench} & 42.0\\
\midrule
\multicolumn{2}{@{}l}{\emph{World--action models}}\\
Fast-WAM \citep{fastwam,memorywam} & 5.9\\
LingBot-VA \citep{lingbotva,memorywam} & 78.2\\
MemoryWAM \citep{memorywam} & 83.0\\
LingBot-VA (time-aware) & 79.9\\
\rowcolor{rtpTableHighlight}
\textbf{RTP (Ours)} & \textbf{84.8}\\
\bottomrule\end{tabular}
\end{table}

\subsection{Ablation Studies}
The RoboMME controls examine learned correction, saved states, action inputs, and historical context.

\textbf{Comparing update policies.}
Fresh-$k$ integrates the complete noise-to-clean path in $k$ intervals; +C adds a correction trained with matched supervision and optimizer updates. Fixed bridge-$k$ uses Bridge-$k$ whenever legal and fresh otherwise. Binary retain/fresh chooses between retention and fresh. Adaptive reconstruction replaces RTP's saved checkpoints with independently re-noised endpoints, retaining the same update choices (Appendix~\ref{app:controls}). RTP has the highest observed success in Table~\ref{tab:core-policy}. Its margins over Fixed bridge-10 and binary retain/fresh are \RoutingGain{} and \BinaryGain{} points; these point estimates characterize the observed success--cost trade-off.

\textbf{History representation.}
Dense sampling uses recent groups; physical coordinates preserve timestamps instead of packed-entry indices. With reset protection fixed, pyramid sampling and physical coordinates yield the strongest Fresh policy (Table~\ref{tab:foundation-ablation}). RTP improves both evaluated configurations, with a larger gain under time-aware history. Each base and bridge is adapted separately.

\begin{table}[!htbp]
\begin{minipage}[t]{.46\linewidth}
\centering\small\setlength{\tabcolsep}{3pt}
\parbox[t][54pt][t]{\linewidth}{\caption{Plan-update policies on RoboMME. Mean call time covers history processing, selection, visual generation, and action decoding.}\label{tab:core-policy}}
\begin{tabular}{@{}LcC@{}}\toprule
Policy & \shortstack{Success\\(\%) $\uparrow$} & \shortstack{Mean call\\(ms) $\downarrow$}\\\midrule
Fresh-20 & 40.4 & 1248\\
Fresh-10+C & 41.8 & 1081\\
Fresh-20+C & 43.5 & 1275\\
\midrule
Adaptive recon. & 45.6 & 1053\\
Fixed bridge-5 & 45.6 & 1057\\
Fixed bridge-10 & 47.4 & 1115\\
Binary retain/fresh & 46.4 & 1113\\
\rowcolor{rtpTableHighlight}
\textbf{RTP (Ours)} & \textbf{48.6} & 1051\\
\bottomrule\end{tabular}

\end{minipage}\hfill
\begin{minipage}[t]{.51\linewidth}
\centering\small\setlength{\tabcolsep}{3pt}
\parbox[t][54pt][t]{\linewidth}{\caption{History ablations on RoboMME. Reset protects the initial observation. RTP gains over matched Fresh use displayed rates.}\label{tab:foundation-ablation}}
\begin{tabular}{@{}LlllC@{}}\toprule
History & Time & Reset & Update & \shortstack{Success\\(\%) $\uparrow$}\\\midrule
Dense & Ordinal & No & Fresh & 33.5\\
Dense & Ordinal & Yes & Fresh & 34.9\\
Dense & Physical & Yes & Fresh & 36.0\\
Pyramid & Ordinal & Yes & Fresh & 37.5\\
Pyramid & Physical & Yes & Fresh & 40.4\\
\midrule
Dense & Ordinal & No & RTP & 37.8\\
\rowcolor{rtpTableHighlight}
Pyramid & Physical & Yes & RTP & \textbf{48.6}\\
\multicolumn{5}{@{}l@{}}{\footnotesize Gain: dense +4.3; time-aware +8.2 pp.}\\
\bottomrule\end{tabular}

\end{minipage}
\end{table}

\begin{figure}[!t]
\centering\includegraphics[width=\linewidth]{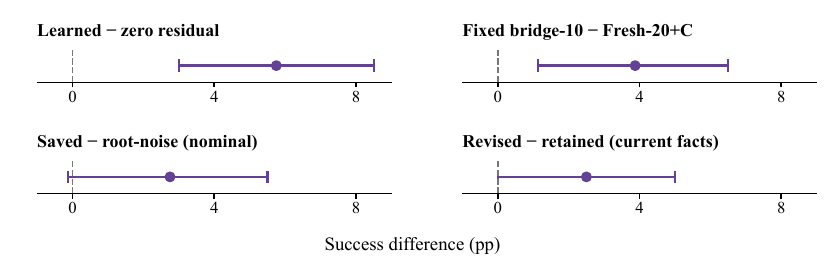}
\caption{\textbf{Paired success differences on RoboMME.} Dots and bars show differences and paired 95\% intervals for the labeled policy contrasts, which are not additive. Sources: Tables~\ref{tab:attribution},~\ref{tab:core-stress}, and~\ref{tab:action-factorial}.}
\label{fig:paired-contrasts}
\end{figure}

\textbf{Learning to revise.}
With source, depth, and action interface fixed, learned correction improves success over zero-residual Fixed bridge-10 by \MainBridgeGain{} points (Table~\ref{tab:bridge-learning}). Fresh-20+C matches feedback, supervision, and optimizer updates and gains \MainExtraLearningGain{} points over Fresh-20; RTP gains a further \MainMatchedGain{} points. Table~\ref{tab:attribution} and Figure~\ref{fig:paired-contrasts} summarize the paired contrasts.

\textbf{Starting from saved states.}
Fixed-depth controls compare saved checkpoints with endpoints re-noised using independent or original-root noise, each with its own fitted correction. Saved checkpoints have higher point estimates, but the nominal source-effect interval spans zero (Table~\ref{tab:core-stress}); their separate contribution remains imprecisely estimated.

\textbf{Action conditioning.}
Table~\ref{tab:action-factorial} varies previous/current facts and retained/revised prefixes at fixed Bridge-10 updating. Under current facts, the prefix effect is \CurrentPrefixGain{} points with interval [\CurrentPrefixLower{}, \CurrentPrefixUpper{}], touching zero. Each cell re-decodes actions and follows its own feedback, so the contrast concerns complete action-conditioning policies.
\begin{table}[H]\centering\small
\caption{Action-input ablation under the Fixed bridge-10 policy. Prefix effects compare revised minus retained using unrounded counts; intervals are paired.}\label{tab:action-factorial}
\begin{tabular}{llccc}\toprule
Action facts & Action prefix & Success (\%) & Prefix effect (pp) & 95\% CI\\\midrule
Previous & Retained & 41.3 & 1.8 & [-0.9, 4.4]\\
Previous & Revised & 43.0 & 1.8 & [-0.9, 4.4]\\
Current & Retained & 44.9 & 2.5 & [0.0, 5.0]\\
\rowcolor{rtpTableHighlight}
Current & Revised & 47.4 & 2.5 & [0.0, 5.0]\\
\bottomrule\end{tabular}\end{table}

\subsection{Computation and Recurrent Behavior}
Adaptive selection allocates generation work across feedback boundaries. RTP averages \RtpMeanSteps{} visual intervals per noninitial call versus \FixedTenMeanSteps{} for Fixed bridge-10, a \RtpStepReduction\% reduction. Their mean full-call times are \RtpMeanCallSeconds{} and \FixedTenMeanCallSeconds{}\,s, respectively (Table~\ref{tab:core-policy}). Full calls also include history processing, selection, and action decoding, so step savings translate into a smaller timing reduction. The 95th-percentile (P95) call time is \RtpPninetyFiveSeconds{} versus \FixedTenPninetyFiveSeconds{}\,s (Appendix~\ref{app:resources}).

Visual work includes fresh calls at exhaustion: a fixed-$k$ policy averages $k+(20-k)f_k$ intervals at fresh fraction $f_k$. Complete-episode recurrent diagnostics track the fraction of selected reusable updates whose offline-expanded visual or action distance exceeds its tolerance. Recurrent fitting lowers this rate relative to the equal-data parent refit, but their success difference has a paired interval spanning zero (Appendix~\ref{app:repeats}).

\subsection{Controlled Feedback Mismatch}
\begin{table}[H]\centering\small
\caption{Fixed-depth source comparisons under actuator holds, 800 keys/condition. I/O/N use independent-noise reconstruction, root-noise reconstruction, and saved checkpoints, respectively, with ten-interval continuation and fresh at exhaustion; difference and paired interval: N minus O.}\label{tab:core-stress}
\begin{tabular}{lccccc}\toprule
Condition & I (\%) & O (\%) & N (\%) & N--O (pp) & 95\% CI\\\midrule
Nominal & 43.6 & 44.6 & 47.4 & 2.8 & [-0.1, 5.5]\\
Four-sample hold & 43.6 & 44.6 & 47.4 & 2.8 & [-0.1, 5.5]\\
Eight-sample hold & 43.0 & 44.2 & 47.1 & 2.9 & [0.0, 5.6]\\
\bottomrule\end{tabular}\end{table}

The early four-sample hold leaves aggregate success unchanged in Table~\ref{tab:core-stress}; the eight-sample hold causes small decreases. The saved-checkpoint advantage over root-noise reconstruction is \NominalSourceGain{} points nominally and \HoldEightSourceGain{} under the longer hold. This setting probes recovery from an early interruption, with substantial time remaining for feedback correction. Figure~\ref{fig:delayed-contact} illustrates a later timing mismatch in the base model, under a separate hold schedule. Further activation-delay and target-shift tests keep nominal parameters frozen (Appendix~\ref{app:stress}). Prediction probes use both a common behavior continuation and each candidate's own executed future, linking visual diagnostics to the actions the candidate induces (Appendix~\ref{app:probes}).

\section{Conclusion}
RTP maintains a visual future as a persistent action condition, revising saved generation states under execution feedback. The learned-residual comparison provides the clearest component evidence; checkpoint-source and prefix effects are less precisely resolved. Removing action supervision improves visual error while lowering success (Table~\ref{tab:loss-terms}), separating prediction accuracy from control utility. Recurrent fitting improves discrepancy control without establishing a task-success gain over the matched refit. Evaluation uses LingBot-VA, finite prediction windows, synchronous execution, and controlled delays. Extending this mechanism across longer interactions offers a direction for persistent, revisable, feedback-driven WAM memory.

\clearpage
\section*{AI Use Statement}
We used generative AI tools to assist with language editing and improve the clarity of the manuscript, as well as to identify relevant literature. The authors take responsibility for the accuracy, originality, and integrity of the final manuscript.

\section*{Reproducibility Statement}
Section~\ref{sec:method} specifies the state representation, revision mechanism, and update policy of RTP. Appendix~\ref{app:training} details the training objectives, hyperparameters, calibration procedure, and checkpoint restoration. Appendices~\ref{app:controls} and~\ref{app:resources} document the evaluation protocol, comparison settings, and computational accounting. Together, these descriptions support implementation and comparison under the stated settings. We will release the code and model checkpoints to the research community.

\section*{Ethics Statement}
This work studies visual-plan revision for robotic control using existing benchmarks and pretrained models. Use of these resources should respect their licenses and access conditions. Because errors in predicted futures can influence executed actions, downstream applications should include task-specific safety evaluation, appropriate operational safeguards, and human oversight suited to the deployment context.

{\small
\bibliographystyle{\colabbibstyle}
\bibliography{refs}
}

\clearpage
\appendix
\section{Training, Calibration and State Restoration}
\label{app:training}\label{app:calibration}\label{app:state}\label{app:distances}
\subsection{Temporal Foundation Configuration}
\label{app:foundation}
\begin{table}[H]\centering\small
\caption{History and model configuration. History-group budgets include all views; task references have a separate budget.}\label{tab:foundation-config}
\begin{tabular}{lcc}\toprule
Quantity & RoboMME & RMBench\\\midrule
Future / consumed groups & $4/1$ & $4/1$\\
Native samples per block $J$ & 4 & 16\\
History budget $B_H$ & 60 & 72\\
Reset anchor quota & 1 & 1\\
Recent nonanchor quota & 12 & 16\\
Task-reference budget $B_M$ & $0$ or $32$ & 0\\
Spatial positions $P$ & 512 & 480\\
Latent channels $C$ & 48 & 48\\
Action / state width & $8/8$ & $16/16$\\
Adaptation updates & 10,000 & 50,000\\
\bottomrule\end{tabular}\end{table}

For current group $n$, spans $q_\ell$, strides $s_\ell$, and reset-fixed phase $i_0=0$, the pyramid candidates are
\begin{equation}
\mathcal K_n=\bigcup_{\ell=1}^{3}
\{i\le n:0\le n-i<q_\ell,\ (i-i_0)\bmod s_\ell=0\}.
\label{eq:pyramid}
\end{equation}
A history group contains one causal latent endpoint for every camera, the action block leading to that endpoint, proprioception, and the original environment-sample endpoint. The reset image is group zero. Groups are the temporal units exposed to the adapted LingBot-VA interface; the raw-observation-to-latent-frame mapping is fixed by preprocessing, and the encoder processes the continuous stream before endpoint selection. $J$ denotes native environment samples represented by one RTP group: RoboMME uses $J=4$ and RMBench uses $J=16$. LingBot-VA's internal chunk parameters are recorded separately; under the released LIBERO configuration \citep{lingbotva,libero}, four latent video positions and four actions per position form one 16-action chunk. Camera identity and spatial positions remain separate from the environment-time coordinate.

The history budgets are fixed at $B_H=60/72$ groups for RoboMME/RMBench, including one reset anchor and at least $12/16$ recent nonanchor groups. A group includes all views; these numbers are neither individual patch tokens nor raw frames. The history pyramid uses $(q_\ell,s_\ell)=(12,1),(76,4),(204,8)$ in group indices. Phase is fixed at the episode reset. On a short record, the selector reserves the anchor, takes up to the recent quota from available nonanchor groups, adds unused pyramid candidates newest first, then fills any remaining budget from all omitted groups newest first. Selection stops at $B_H$, and the selected records are sorted by physical endpoint. Thus a short trajectory uses every available group and a long trajectory never exceeds its budget.

RoboMME task-reference clips are available context, not current-episode observations. Reference sampling selects at most 125 frames at uniformly spaced original reference indices, retaining endpoints; the first frame and subsequent blocks of four produce at most 32 reference groups. The original reference indices retain an independent reference coordinate system, with padding masked in an incomplete final block. There is no environment reset anchor inside this stream. Tasks without a reference use $B_M=0$; reference tasks use $B_M=32$. RMBench uses its task instruction and $B_M=0$. The language instruction is always separate from the group budgets.

Each Stage-A training item uses the pyramid selector with probability 0.8 or an anchor-plus-most-recent selector with probability 0.2, with the same $B_H$. Targets always comprise the next $H$ contiguous groups and their aligned actions. Deployment uses the specified pyramid selector. Factor controls use separately trained bases: the dense-history comparator uses the latest $B_H$ groups, packed ordinal positions, and no protected reset anchor; the temporal base uses the full construction above. Both use the same pretrained initialization, training manifest, reference policy, number of updates, and physical action convention. The dense training branch of the temporal model is not the separately trained dense comparator.

Factual positions use the current-boundary coordinate $p_i^{\rm fact}=(\rho_i-b_n)/J$, while a saved predictive plan retains $p_t^{\rm plan}=(\tau_t-b_{\rm root})/J$. An accepted prefix is re-expressed relative to the current boundary for the action query without changing the saved solver root. The reference stream uses its own sample index divided by its encoding block size and a distinct sequence identifier. Each invocation rebuilds factual and query positions and masks, so the same factual record can be presented in the coordinate view required by its consumer.

The causal observation encoder never reads a predicted frame. Factual-context processing cannot attend to predictive tokens. Visual queries can read all selected facts through the current boundary and use the backbone's predictive-window mask. The action branch reads current factual context and the accepted clean prefix. No generated consumed position is reclassified as evidence.

\subsection{Data, Supervision and Fitting}
\label{app:fitting}
The temporal foundation comprises the adapted model and its history conventions. Its signature records weights, normalization, history sampling, anchor/reference budgets, positions, masks, and action coordinates. This signature is shared only within a matched history configuration and fitting run. Prefix and suffix tensors have shapes $H_c\times P\times C$ and $(L_n-H_c)\times P\times C$, with $P$ spatial positions across views and $C$ latent channels.
Table~\ref{tab:foundation-ablation} fixes the factor cells. The anchored $2\times2$ holds one protected reset group, recent quotas, references, and total capacity fixed while varying the sampling mixture and ordinal versus physical positions. Dense sampling uses the anchor-plus-latest selector at every training/deployment call, whereas pyramid sampling uses the specified training mixture and pyramid deployment. The unanchored dense row and anchored dense/ordinal row isolate anchor protection at equal capacity. The interaction contrast compares Dense+RTP with its own Fresh and Temporal+RTP with its own Fresh; each base signature has its own bridge.

Stage A adapts the video--action model while freezing its pretrained observation and language encoders. In native normalized coordinates, its conditional flow objectives are
\begin{equation}
\begin{aligned}
z_u&=(1-u)\epsilon_v+uZ,\qquad a_w=(1-w)\epsilon_a+wA,\\
\mathcal L_{\rm base}
&=\mathbb E\!\left[
\|v_\theta(z_u,u,G,M)-(Z-\epsilon_v)\|_{\rm av}^2
+\|w_\phi(a_w,w,G,\operatorname{Prefix}_{H_c}(Z),M)-(A-\epsilon_a)\|_{\rm av}^2
\right].
\end{aligned}
\label{eq:base-loss}
\end{equation}
$\epsilon_v,\epsilon_a$ are independent standard Gaussian tensors; $u,w$ are independently sampled on $(0,1)$. Each squared norm averages valid entries of its modality, so camera resolution does not set the action-loss weight. The first $H_c$ groups of the clean future $Z$ teacher-condition the next $J$ action samples during adaptation. At deployment the corresponding argument is the generated prefix. The two branches retain the pretrained video--action interface and share the adapted backbone where defined by that architecture.

Base adaptation uses only benchmark training demonstrations, with RMBench's 50 demonstrations/task and RoboMME's supplied training manifest. AdamW uses learning rate $10^{-5}$, coefficients $(0.9,0.95)$, weight decay 0.1, effective batch 8, ten linear warmup updates, and a constant rate thereafter. The final checkpoint follows 10,000 RoboMME or 50,000 RMBench updates. The foundation signature contains dataset digests, training seed, sampling configuration, preprocessing, and model checkpoint digest.

With the foundation frozen, trajectory collection uses the fresh controller on disjoint reset keys and retains success and failure episodes. Collection trajectories have zero-based indices within each task: indices congruent to 4 modulo 5 form calibration, all others fitting. Collection starts with 100 trajectories/task and extends in blocks of 20 until each split contains its required valid tuples. A tuple needs an archived fresh parent, complete applied blocks through its selected feedback endpoint, a complete following action block, and a nonempty subsequent observation intersection. Recording ends at terminal flags. Tuples are uniquely keyed by trajectory, parent root, and feedback boundary; trajectory-level splitting prevents overlap between fitting and calibration.

Fitting-set sampling selects 800 tuples/task without replacement using seed 1: 12,800 on RoboMME and 7,200 on RMBench. Calibration sampling uses seed 0: 256/task on RoboMME (4,096 total), and 456 for RMBench R01 plus 455 for each remaining task (4,096 total). Sampling follows a fixed collection manifest. All tuples from a trajectory stay in one split. An additional, separate cost-matching pool contains 20 trajectories/task for cost-matching calibration. Diagnostic and test resets occur in none of these pools.

Each fresh parent window is archived while the behavior controller continues to replan from current facts. Stage-B tuples sample feedback at consumed indices $c=1,2,3$ of that archived window, giving remaining lengths $3,2,1$ and source ages $1,2,3$. The tuple descriptor aligns the archived prediction at the selected feedback endpoint with the observation there. Within each task, the 800 fitting tuples use feedback-boundary quotas $(267,267,266)$. Calibration quotas are $(86,85,85)$ for RoboMME, $(152,152,152)$ for RMBench R01, and $(152,152,151)$ for its other tasks. Collection continues until every quota is filled; parent identity and original checkpoint metadata remain fixed. The fresh behavior continuation from the selected physical state supplies observed future groups and the next action target. The action target is the decoder's continuous denormalized command before actuator clipping/discrete gripper conversion; factual records contain the separately recorded applied commands.

Default fresh-parent bridge fitting jointly optimizes the residual and DiffEncode through both depths. An independent fresh reference root remains fixed per tuple and epoch; revised and reference calls within that tuple share an action seed. Bridge and estimator fitting both use AdamW, learning rate $10^{-4}$, coefficients $(0.9,0.95)$, weight decay $10^{-2}$, effective batch 16, ten epochs, gradient-norm clipping at 1, and the final epoch. Fitting uses microbatch one with accumulation, activation recomputation, and float32 reductions. Frozen base parameters have gradients disabled, but their operations remain differentiable with respect to the residual-dependent inputs.

Per tuple, two-depth bridge fitting executes 15 differentiable visual intervals and two 50-interval action calls; two-root Fresh-20+C executes 40 visual intervals with the same action work. Reference construction is separate. Expanding retain, both bridges and two fresh references costs at most 55 visual and 250 action intervals. Forward counts include every executed call and count shared features once. The reported training-work counts separate tuple expansion from differentiable fitting; Appendix~\ref{app:resources} reports controller timing, persistent storage, and parameter counts.

Stage C freezes the fitted bridge before producing fitting and calibration labels. Each tuple has two independently indexed fresh visual roots and a common action seed across retain, both bridges, and both references. Each label is associated with foundation, bridge, tuple, mode, root, and seed identifiers. Calibration estimates no parameters of the bridge or backbone.

\subsection{Distances and Trainable Modules}
For clean latent coordinates $z_{tpc}$ and continuous denormalized action coordinates $A_{jd}$,
\begin{equation}
d_v(z,z';I)=\frac{1}{|I|PC}\sum_{t\in I,p,c}
\frac{(z_{tpc}-z'_{tpc})^2}{(\sigma_c^v)^2},\qquad
d_a(A,A')=\frac{1}{JD_a}\sum_{j,d}
\frac{(A_{jd}-A'_{jd})^2}{(\sigma_d^a)^2}.
\label{eq:distances}
\end{equation}
Scales are estimated only from Stage-B fitting targets, with standard deviations floored at $10^{-6}$ and then frozen. The distance includes continuous gripper coordinates and excludes padding. Every action branch uses the same native coordinate conventions and normalization inverse; physical clipping and gripper conversion happen after decoding for execution. Supervision propagates through continuous values before those discrete operations.

Predictions and observations use the same clean-latent coordinate system. Nominal groups have identical endpoint grids. Variable-grid diagnostics use linear interpolation within the observed overlap and average valid coordinates, with support confined to observed endpoints from the same camera. The teacher term uses recorded future intersection $I_*$, while fresh consistency uses the candidate/fresh intersection $I_T$. For $H=4$ and consumed index $c$, supervision covers the remaining $4-c$ groups. Later fresh-reference groups outside that old window are masked.

Action re-decoding is an independent conditional flow solve. The solve initializes normalized action noise $a_0$ from the indexed action seed and spans $N_a=50$ intervals:
\begin{equation}
\begin{aligned}
a_{l+1}&=a_l+(w_{l+1}-w_l)w_\phi(a_l,w_l,G,P,M),\\
A&=\operatorname{Denorm}(a_{N_a}),\qquad
\frac{\partial\mathcal L_a}{\partial\eta}
=\frac{\partial\mathcal L_a}{\partial A}
\frac{\partial\pi_\phi}{\partial P}
\frac{\partial P}{\partial\eta}.
\end{aligned}
\label{eq:action-flow}
\end{equation}
Here $\mathcal L_a=d_a(A,A^*)$. The time grid uses the benchmark's action shift. The action field receives current facts and the accepted prefix, with no trainable RTP residual on the action field itself. Gradients through the complete decoder and resumed visual steps train the visual correction; disabling parameter gradients is distinct from wrapping these calls in a no-gradient execution context.

The pooling configuration is as follows. Spatial/view-average latent groups have width $C=48$. Ground and instruction summaries are 1536-wide fixed pair-averages of their 3072-wide projected backbone tokens, averaged over valid tokens; this adds no trained projection. An absent visual reference leaves the instruction summary available. DiffEncode concatenates three latent vectors (prediction, observation, difference), flattened applied controls, three proprioceptive vectors (old, new, difference), and the ground/task summaries:
\begin{equation}
D_\Delta=3C+JD_a+3D_s+2D,\qquad
D=1536,\quad(D_a,D_s)=(8,8)\ {\rm or}\ (16,16).
\label{eq:delta-width}
\end{equation}
Thus $D_\Delta=\RoboDeltaWidth{}$ on RoboMME and $\RmbDeltaWidth{}$ on RMBench. Its MLP has widths $D_\Delta\to D\to D$, with GELU after the first layer.

The backbone visual width is $D_b=3072$, latent channels $C=48$, and patch size $(1,2,2)$, so $C_p=4C=192$. The representation $h_\theta$ is taken after final normalization and time modulation, immediately before the visual output projection. With first active index $c$,
\begin{equation}
c_\eta=\operatorname{Linear}_{1538\to1536}([\Delta,u,c/H]),\qquad
r_\eta=\mathcal U\!\left(
\operatorname{MLP}_{4608\to1536\to192}([h_\theta,c_\eta])\right).
\label{eq:residual-shape}
\end{equation}
The conditioning vector $c_\eta$ is broadcast over visual tokens. The residual uses GELU between its two linear layers and zero-initializes the final weights and bias. For batch $B_{\rm tr}$ and $L_{\rm tok}=HP/4$, shapes are $B_{\rm tr}\times L_{\rm tok}\times4608$, then $B_{\rm tr}\times L_{\rm tok}\times192$. $\mathcal U$ restores $B_{\rm tr}\times H\times P\times48$ in native temporal/view/spatial order. It does not mix frames or views.

Source age is candidate-specific: the difference between the current physical boundary and the saved state's immutable creation boundary, divided by $J$, gives age in groups; division by $H$ gives its input feature. Retain uses the clean endpoint's boundary; each native revision depth uses its own checkpoint's boundary. Reconstructed candidates inherit the endpoint's source boundary, not the time at which interpolation is computed. Updating a record binding never resets these ages. A bridge stamps its new endpoint and newly visited checkpoints with the current boundary; its restored starting checkpoint keeps its old boundary. Root age and remaining-window length are different quantities.

The estimator concatenates $\Delta$ and the ground summary, a 48-wide active-plan mean, 48-wide checkpoint mean and standard deviation, and three scalars: remaining length/$H$, remaining solver steps/$N$, and source age/$H$. Retain uses the clean full-window statistics in place of a checkpoint. This gives $D_S=2D+3C+3=\SelectorWidth{}$. Two hidden layers of width $D_S$ with GELU feed four outputs: two softplus locations and two softplus scales plus $10^{-6}$. The estimator objective is the tuple average of
\begin{equation}
\mathcal L_S=\sum_{m\in\mathcal V_n,r\in\{v,a\}}
\left[\frac{|y_{r,n}^m-\widehat d_{r,n}^m|}
{\widehat u_{r,n}^m}+\log\widehat u_{r,n}^m\right].
\label{eq:estimator-loss}
\end{equation}
This is a heteroscedastic absolute-error objective. The scale estimates discrepancy-estimation error.

Disjoint calibration tuples determine
\begin{equation}
\tau_r=Q_{.90}\{d_r(T_1,T_2)\},\qquad
\beta=\max\left(0,Q_{.95}
\left\{\max_{m\in\mathcal V_n,r}
\frac{y_{r,n}^m-\widehat d_{r,n}^m}
{\widehat u_{r,n}^m}\right\}\right).
\label{eq:calibration}
\end{equation}
Quantiles use the nearest rank, sorted index $\lceil pK\rceil$. The visual $T_1$--$T_2$ distance uses the same unexecuted old-window timestamp intersection as candidate labels. Calibration is specific to each benchmark and fitted base. For the primary runs, $(\tau_v,\tau_a,\beta)$ is $(\RoboTauV{},\RoboTauA{},\RoboBeta{})$ on RoboMME and $(\RmbTauV{},\RmbTauA{},\RmbBeta{})$ on RMBench; selection uses their unrounded values. The primary RTP thresholds remain unchanged at evaluation. These empirical quantiles describe the feedback-boundary-stratified archived-parent calibration distribution and define fresh-relative discrepancy tolerances on that support. Recurrent fitting is evaluated separately for later states created by accepted updates; the correction modules are refitted from their original initialization on recurrent or matched archived-parent tuples, followed by their discrepancy estimators. The estimation-error margin is recalibrated while the primary distance tolerances remain fixed (Appendix~\ref{app:repeats}). Separate diagnostics characterize later-boundary and perturbed-state reliability.

\subsection{Native Windows and Checkpoints}
The complete persistent record is $Q_n=(x_N,\mathcal S_n,b_{\rm root},c_n,g_n,\ell_n)$: $g_n$ identifies the current factual binding and $\ell_n$ the root and its indexed random seed; the other fields are defined in Section~\ref{sec:foundation}. Available checkpoints form $\mathcal S_n$. Each $S_{n,j}$ retains the immutable provenance of the solver state it stores.
The solver uses noise-to-clean coordinates with visual shift $\gamma=5$:
\begin{equation}
u_j=\frac{j/N}{\gamma-(\gamma-1)j/N},\quad j=0,\ldots,N,
\qquad h_j=u_{j+1}-u_j.
\label{eq:schedule}
\end{equation}
A native implementation exposing clean-to-noise time uses the corresponding coordinate reversal and velocity sign. Checkpoints store states immediately before intervals 10 and 15. A five-step branch restores $x_{15}$ and traverses intervals 15--19; a ten-step branch restores $x_{10}$ and traverses 10--19. Actual network-evaluation counts include native guidance call multiplicity.

A checkpoint consists of the complete latent tensor, solver index/time, root origin, camera/spatial layout, normalization signature, root RNG identifier, and immutable source fact version. Raw latents are saved, not positionalized attention keys. Each restoration first rebuilds the selected factual interface and its fact/query positions and masks under current facts, then recomputes the attention keys and values used by the resumed solver.

Retain preserves full-window clean contents and checkpoints, updating only the active index and current fact binding. Each bridge preserves the accepted continuation, its immutable starting checkpoint with its old source version, and newly visited required checkpoints with the current version; older ancestor records are excluded. Bridge-5 therefore has only $x_{15}$; Bridge-10 has $x_{10}$ and its newly visited $x_{15}$; fresh has both from its new root. The stored corrected endpoint covers the full window, including consumed positions; prefix extraction reads only active positions.

\textbf{Structural invariants.}
For the action call, re-encoding means applying the native patch embedding and current-origin positional encoding to the accepted clean latent prefix. It does not mean decoding to pixels and passing those pixels through the observation encoder. No old predictive attention cache is reused.

Initially, facts contain only reset observations and the prediction record contains exactly $H$ future groups. History updates append only observed/applied records, so induction preserves factual purity. Every residual output has the full solver shape, so each Euler update is well typed. Execution advances the consumed index once to $c_n^+=c_n+H_c$; reuse stores this index and requires at least $H_c$ remaining groups; hence a root permits at most $\lfloor H/H_c\rfloor-1$ reusable feedback updates. Fresh restores $c=0$, and the empty legal set always falls back to fresh. With fixed tie-breaking, the update is defined at every complete nonterminal boundary. These invariants concern state consistency, independently of task success.

\section{Common Evaluation and Complete-Policy Comparisons}
\label{app:controls}\label{app:comparison-protocol}\label{app:comparison-results}
\subsection{Execution and Statistical Units}
RoboMME uses two $256\times256$ views, eight continuous action coordinates, and eight proprioceptive coordinates. RMBench uses front $256\times320$ and two wrist $128\times160$ views, with sixteen action/proprioceptive coordinates. Latent spatial stride 16 gives $P=512$ and $480$ across cameras. Visual/action guidance is 5/1, visual generation uses 20 intervals, and action generation uses 50. The action noise-to-clean schedule uses Equation~\ref{eq:schedule} with its own interval count and shift 0.05/1 for RoboMME/RMBench. Within each matched history configuration and fitting run, internal policies share adapted weights, masks, scales, and the action interface.

Primary evaluation alternates a complete controller call and execution and ends at the first terminal flag. Evaluation records contain both issued and applied control streams. Visual roots are indexed by a deterministic hash of benchmark, fitting seed, reset key, rollout seed, physical boundary, and solver role. Compared candidates at a common boundary share the action-noise index. Roots remain independent across roles. Whole policies share reset and initial randomness, then act on their own observations.

For task $t$ and reset $i$, $Y^F_{ti},Y^R_{ti}$ denote binary outcomes. Equal task sizes make task-averaged and pooled success identical; unequal diagnostic eligibility does not. For $K$ tasks and $n_t$ resets/task,
\begin{equation}
S^m=\frac{100}{K}\sum_t\frac{1}{n_t}\sum_iY^m_{ti},
\qquad
\Delta S=\frac{100}{K}\sum_t\frac{1}{n_t}
\sum_i(Y^R_{ti}-Y^F_{ti}).
\label{eq:success}
\end{equation}
Repeated rollout seeds are averaged within each key. Intervals use 10,000 within-task bootstrap resamples of keys with seed 20260917, keeping all modes and repeats paired, and take the nearest-rank 2.5th/97.5th percentiles. The independent sampling unit is the reset key. These intervals are conditional on the fitted weights. Comparisons without a paired interval are descriptive point estimates; an unreported interval is not evidence of either significance or equivalence.

Let $F,R$ denote the two success counts among $N$ shared reset keys, and let $d$ count fresh-only successes. The margins imply, but do not select, a paired matrix:
\begin{equation}
\begin{aligned}
(n_{11},n_{10},n_{01},n_{00})&=(F-d,\ d,\ R-F+d,\ N-R-d),\\
\max(0,F-R)&\le d\le\min(F,N-R).
\end{aligned}
\label{eq:paired-margins}
\end{equation}
All entries are integers. The ordering is both success, fresh only, RTP only, neither. Rescues count RTP-only successes ($n_{01}$); regressions count fresh-only successes ($n_{10}$). Tables~\ref{tab:tasks}--\ref{tab:rmb-tasks} report per-task success and discordant-pair counts. The pooled RMBench difference is $\RmbNetPoints{}$ percentage points.

\begin{table}[!htbp]\centering\small
\caption{Per-task RoboMME comparisons, 50 paired keys/task. Successes, rescues, and regressions are counts; differences and intervals are in percentage points. -- denotes unreported paired statistics.}\label{tab:tasks}
\begin{tabular}{lccccc}
\toprule Task & Fresh & RTP & Rescues & Regressions & $\Delta$ [95\% CI]\\
\midrule
T01 & 13 & 13 & -- & -- & 0.00 [--]\\
T02 & 12 & 16 & 7 & 3 & 8.00 [-4.0, 20.0]\\
T03 & 11 & 14 & 4 & 1 & 6.00 [-2.0, 14.0]\\
T04 & 19 & 24 & 9 & 4 & 10.00 [-4.0, 24.0]\\
T05 & 16 & 18 & 5 & 3 & 4.00 [-6.0, 16.0]\\
T06 & 23 & 33 & 13 & 3 & 20.00 [6.0, 34.0]\\
T07 & 24 & 23 & 5 & 6 & -2.00 [-14.0, 10.0]\\
T08 & 22 & 29 & 8 & 1 & 14.00 [4.0, 26.0]\\
T09 & 17 & 24 & 8 & 1 & 14.00 [4.0, 26.0]\\
T10 & 16 & 21 & 8 & 3 & 10.00 [-2.0, 22.0]\\
T11 & 21 & 27 & 7 & 1 & 12.00 [2.0, 22.0]\\
T12 & 20 & 23 & 5 & 2 & 6.00 [-4.0, 16.0]\\
T13 & 29 & 30 & 5 & 4 & 2.00 [-10.0, 14.0]\\
T14 & 28 & 31 & 5 & 2 & 6.00 [-4.0, 16.0]\\
T15 & 27 & 31 & 7 & 3 & 8.00 [-4.0, 20.0]\\
T16 & 25 & 32 & 8 & 1 & 14.00 [4.0, 26.0]\\
\bottomrule
\end{tabular}
\end{table}

\begin{table}[H]\centering\small
\caption{Per-task RMBench comparisons, 100 paired keys/task. Rescues and regressions are counts; differences and intervals are percentage points. -- denotes unreported paired statistics.}\label{tab:rmb-tasks}
\begin{tabular}{lccccc}\toprule
Task & Fresh & RTP & Rescues & Regressions & $\Delta$ [95\% CI]\\\midrule
R01 & 61 & 63 & -- & -- & 2.00 [--]\\
R02 & 79 & 83 & 9 & 5 & 4.00 [-3.0, 11.0]\\
R03 & 77 & 81 & 6 & 2 & 4.00 [-1.0, 10.0]\\
R04 & 73 & 82 & 11 & 2 & 9.00 [2.0, 16.0]\\
R05 & 79 & 88 & 10 & 1 & 9.00 [3.0, 15.0]\\
R06 & 80 & 86 & 8 & 2 & 6.00 [0.0, 12.0]\\
R07 & 89 & 92 & 5 & 2 & 3.00 [-2.0, 8.0]\\
R08 & 93 & 95 & 3 & 1 & 2.00 [-2.0, 6.0]\\
R09 & 88 & 93 & 8 & 3 & 5.00 [-1.0, 12.0]\\
\bottomrule\end{tabular}\end{table}

\begin{table}[H]\centering\small
\caption{Paired policy and intervention contrasts, in percentage points. Overlapping contrasts are not additive module contributions. -- denotes unreported intervals.}\label{tab:attribution}
\begin{tabular}{lcc}\toprule
Contrast & Difference & 95\% interval\\\midrule
Foundation configuration & 6.88 & [4.13, 9.63]\\
Extra feedback fitting & 3.12 & [0.50, 5.63]\\
RTP vs. matched Fresh-20+C & 5.125 & --\\
Learned vs. zero, Fixed bridge-10 & 5.75 & [3.00, 8.50]\\
RTP vs. Fixed bridge-10 & 1.250 & --\\
RTP vs. binary retain/fresh & 2.250 & --\\
Fixed bridge-10 vs. matched Fresh-20+C & \BridgeMatchedGain{} & [\BridgeMatchedLower{}, \BridgeMatchedUpper{}]\\
\bottomrule\end{tabular}\end{table}

T01--T16 follow benchmark-manifest order within the four named suites, and R01--R09 follow RMBench-manifest order. Evaluation outputs associate these identifiers with exact task names and the manifest digest. RoboMME entries for $\pi_{0.5}$, MemER, and MME-VLA follow the benchmark configurations in \citet{robomme}. RMBench entries for $\pi_{0.5}$, X-VLA, and Mem-0 follow \citet{rmbench}; Fast-WAM, LingBot-VA, and MemoryWAM follow \citet{memorywam}. The dense-history and time-aware LingBot-VA variants are separately finetuned under their respective history configurations.

\subsection{Matched Fresh and Reconstructed Controllers}
Fresh-$k$+C consumes the same feedback descriptor and uses the residual/DiffEncode architecture, but sets its generation starting state to new Gaussian noise. Each complete fresh schedule with $k\in\{5,10,12,20\}$ intervals has a separately trained correction. Each uses the primary tuple keys, targets, two-root action supervision, optimizer updates, and final-epoch rule. Supervision covers the same $4-c$ remaining old-window groups, with later fresh groups masked; the fresh correction receives the matched parent's $c/H$ in its conditioning vector. Equation~\ref{eq:schedule} with the chosen interval count spans the whole path. For each tuple, the same three-term bridge objective is averaged over two independently seeded full-path fresh candidates in place of the two continuation depths, using an independent frozen fresh reference. Shared initial generation uses the uncorrected base because no feedback descriptor exists yet. This matches trainable architecture, supervision and optimizer updates, not the number of differentiable visual intervals.
\begin{table}[H]\centering\small
\caption{Fresh integration schedules and RTP. Visual work includes structural refreshes; differences are RTP minus row.}\label{tab:fresh-frontier}
\begin{tabular}{lcccc}\toprule
Policy & Visual steps & Success (\%) & Call (s) & RTP diff. (pp)\\\midrule
Fresh-5 & 5.00 & 34.25 & 0.976 & 14.38\\
Fresh-5+C & 5.00 & 36.50 & 0.991 & 12.13\\
Fresh-10 & 10.00 & 38.12 & 1.066 & 10.50\\
Fresh-10+C & 10.00 & 41.75 & 1.081 & 6.88\\
Fresh-12+C & 12.00 & 42.50 & 1.123 & 6.13\\
Fresh-20+C & 20.00 & 43.50 & 1.275 & 5.13\\
\rowcolor{rtpTableHighlight}
RTP & 7.78 & 48.63 & 1.051 & 0.00\\
\bottomrule\end{tabular}\end{table}

Independent reconstruction forms $x_u^{I}=(1-u)\epsilon'+ux_N$ at the native revision time, using the current complete clean endpoint and independently indexed noise. Original-root reconstruction forms $x_u^{O}=(1-u)\epsilon_\ell+ux_N$, using the ancestor root. Each source receives its own matched residual/DiffEncode and corresponding estimator labels. Fixed reconstruction-10 uses ten continuation intervals whenever enough of the active plan remains and fresh at exhaustion. Adaptive reconstruction and Adaptive root-noise reconstruction allow both reconstructed depths, using independent noise and the original root noise, respectively; both also include retention and fresh.

A native checkpoint need not lie on the root--endpoint chord. For a twice-differentiable path with $\sup_{s\in[0,1]}\|x''(s)\|\le M$, the interpolation remainder satisfies
\begin{equation}
\|x(u)-[(1-u)x(0)+ux(1)]\|\le\tfrac12 M u(1-u).
\label{eq:chord}
\end{equation}
The zero-endpoint Green-kernel representation gives the bound by integrating its absolute mass. For $x(u)=e^u$, the midpoint is $e^{1/2}$ rather than $(1+e)/2$; straight paths give equality at zero error. The statement distinguishes state construction and makes no ordering claim about control success. Discrete solver error is an additional difference.

The zero-residual control uses the same native-checkpoint construction and current facts while removing the learned velocity correction (Table~\ref{tab:bridge-learning}). The source comparisons isolate native versus reconstructed states at matched continuation depth.

Cost-matched adaptive reconstruction policies use one positive multiplier on both tolerances, fitted only on the separate nominal cost pool. The search covers a fixed 41-point log-multiplier grid from $-2$ to $2$ and minimizes the absolute mean-step difference from primary RTP, breaking ties toward lower cost and then smaller multiplier. The achieved difference characterizes the match. The multiplier remains frozen during test and perturbation evaluation; a matching target does not guarantee matching off-distribution cost.
Binary retain/fresh uses the primary retain heads and unchanged thresholds. It retains the unexecuted plan when valid and both scores pass, and selects fresh otherwise. It changes the candidate family without fitting another success predictor. On any identical pre-decision record, binary and full RTP must agree on whether to retain: both test the same mode-zero score first. Shared reset and initial randomness make this equality mandatory at their first feedback boundary. After their actions diverge, decision-level equivalence applies only at identical states; aggregate retain fractions need not remain equal.

\section{Action Attribution, Update Selection and Recurrent Support}
\label{app:routing}\label{app:repeats}
\subsection{Action Inputs}
Every factorial cell updates the visual record with Bridge-10 under current facts. The action call independently uses previous/current factual content and retained/revised active prefix. Both factual inputs are positioned at the current action origin; previous facts stop at the preceding boundary and contain no new observations. Action noise is common within each intervention. Structural fresh fallback uses current facts in all cells. The visual record stores the revised plan in every cell, even if the action decoder reads the retained prefix. All cells subsequently follow their own actions and feedback.

\subsection{Selection Controls}
Fixed bridge-5 and Fixed bridge-10 always select their named native continuation when an action-sized unexecuted prefix remains and fresh at exhaustion. They share the same learned bridge, factual interface, and action decoder as RTP. Binary retain/fresh keeps the primary retain heads and thresholds but removes both bridge candidates. Table~\ref{tab:routing-budget} compares these simple policies with full RTP, including forced refreshes and each policy's own stopping time.

The primary candidates are retention, five-step revision, ten-step revision, and fresh-20. This is an interface-defined candidate set, not a claim that these budgets are optimal. A fifteen-step extension needs the native $x_5$ checkpoint, corresponding bridge supervision, discrepancy labels, calibration, and a complete-policy evaluation. Its performance is a separate experimental question from the evaluated depths.

Calibration and cost matching use only reserved nominal data. Selected thresholds remain frozen during testing, and cost multipliers depend exclusively on the reserved cost pool. Comparisons jointly characterize success and full-call cost, including the accuracy and cost of fixed-depth policies.

\begin{table}[H]\centering\small
\caption{Fixed-depth, binary retain/fresh, and adaptive updates. Each policy includes exhaustion refreshes.}\label{tab:routing-budget}
\begin{tabular}{lccc}\toprule
Policy & Success (\%) & Mean steps & Call (s)\\\midrule
Fixed bridge-5 & 45.62 & 8.49 & 1.057\\
Fixed bridge-10 & 47.38 & 12.33 & 1.115\\
Binary retain/fresh & 46.38 & 11.10 & 1.113\\
\rowcolor{rtpTableHighlight}
RTP & 48.63 & 7.78 & 1.051\\
\bottomrule\end{tabular}\end{table}

\subsection{Recurrent Fitting and Diagnostics}
Default training covers all three reusable feedback boundaries of archived fresh parents. The age/length ablation jointly removes those lifecycle coordinates and refits the estimator with adjusted layer widths on the same tuples. Recurrent fitting uses a disjoint pool collected under the frozen default policy and its accepted records, including newly created endpoints and missing checkpoints. Valid tuples follow the same per-task feedback-boundary quotas as default training. A fresh behavior continuation cloned from each physical state supplies targets. Bridge fitting averages the loss in Section~\ref{sec:revision} over the depths with available checkpoints in each tuple. The matched parent refit uses a separate archived-parent pool with identical tuple counts, feedback-boundary quotas, epochs, and optimizer settings. Within each task and feedback-boundary stratum, it mirrors the recurrent depth-availability masks for correction fitting, estimator fitting, and calibration. Both corrections are fitted from their original initialization, followed by their estimators. All rows use the primary distance normalization and frozen tolerances; only the estimation-error multiplier is recalibrated on a disjoint, feedback-boundary-matched split. Table~\ref{tab:recurrent-controls} reports whole-policy success alongside the discrepancy checks.
The tolerance-exceedance rate divides selected reusable updates whose offline-expanded visual or action distance exceeds its frozen tolerance by all selected reusable updates.
\begin{table}[H]\centering\small
\caption{Recurrent discrepancy checks on 320 episodes/condition, with common frozen tolerances. Entries are exceedances / selected reuse (percent).}\label{tab:recurrent-calibration}
\begin{tabular}{lccc}\toprule
Fitting & Nominal & Hold 4 & Hold 8\\\midrule
Default RTP & 69/4822 (1.43\%) & 91/4542 (2.00\%) & 94/4252 (2.21\%)\\
Without age/length & 181/4726 (3.83\%) & 176/4392 (4.01\%) & 219/4141 (5.29\%)\\
Matched parent refit & 74/4838 (1.53\%) & 91/4533 (2.01\%) & 93/4257 (2.18\%)\\
Recurrent fitting & 24/4944 (0.49\%) & 25/4666 (0.54\%) & 24/4376 (0.55\%)\\
\bottomrule\end{tabular}\end{table}

Under nominal conditions, recurrent fitting minus the matched parent refit changes the tolerance-exceedance rate by \RecurrentExceedanceGain{} percentage points (paired episode-bootstrap 95\% interval [\RecurrentExceedanceLower{}, \RecurrentExceedanceUpper{}]). These rates condition on each policy's selected reuse; whole-policy outcomes use every reset.
\begin{table}[H]\centering\small
\caption{Complete-policy recurrent-fitting controls on 800 reset keys. The first four differences are row minus default RTP; the final row is their direct recurrent-minus-matched contrast, not an additional policy. Differences and intervals are in percentage points; -- denotes unreported intervals.}\label{tab:recurrent-controls}
\begin{tabular}{lccc}\toprule
Policy & Success (\%) & Mean steps & Difference [95\% CI]\\\midrule
Default RTP & 48.63 & 7.78 & 0.00 [0.00, 0.00]\\
Without age/length & 48.00 & 7.80 & -0.625 [--]\\
Matched parent refit & 49.38 & 7.79 & 0.750 [--]\\
Recurrent fitting & 49.50 & 7.75 & 0.875 [--]\\
\midrule
Recurrent $-$ matched refit & -- & -- & \RecurrentMatchedGain{} [\RecurrentMatchedLower{}, \RecurrentMatchedUpper{}]\\
\bottomrule\end{tabular}\end{table}

The direct recurrent-minus-parent success contrast is \RecurrentMatchedGain{} percentage points (95\% interval [\RecurrentMatchedLower{}, \RecurrentMatchedUpper{}]). Improved discrepancy control therefore does not establish a corresponding task-success improvement over the matched refit. The discrepancy diagnostic and success endpoint use their respective episode pools and denominators.

For a diagnostic pool of 320 episodes, let $R_j,U_j,F_j$ be reached, selected-reuse, and fresh decisions at root feedback boundary $j$. Let $T_0$ count roots terminating before first feedback and $T_j$ roots terminating after reuse at boundary $j$. If $K_{\rm root}$ includes initialization and every fresh root, then
\begin{equation}
\begin{aligned}
R_j&=U_j+F_j,\quad U_4=0,\quad
K_{\rm root}=320+\sum_{j=1}^{4}F_j,\\
R_1&=K_{\rm root}-T_0,\qquad
R_{j+1}=U_j-T_j\ (j=1,2,3),\qquad
T_0+\sum_{j=1}^{3}T_j=320.
\end{aligned}
\label{eq:lifecycle-counts}
\end{equation}
These identities accommodate early fresh, missing checkpoints, and episode termination. Candidate-count totals additionally sum each reached state's legal set size and are not determined by root counts.
Each reusable candidate has a pre-selection record containing estimated discrepancies, discrepancy-estimation error scales, offline labels, and checkpoint availability for Equation~\ref{eq:route}. Binary and full policies share retain scores at identical states. A false-fresh decision selects fresh despite at least one legal reusable candidate satisfying both tolerances in its offline labels. Exceedances and false-fresh decisions use these discrepancy labels; success labels never enter update selection. Table~\ref{tab:diagnostic-details} and the default nominal calibration cell use the same decision-record definitions.
\begin{table}[H]\centering\small
\caption{Diagnostics by feedback boundary within each root. Exceedances count selected reuse with either distance above tolerance. N/A denotes the undefined zero-reuse ratio $0/0$.}\label{tab:diagnostic-details}
\begin{tabular}{cccccc}\toprule
Boundary & Reached & Reused & Fresh & Exceedances & Exceedance (\%)\\\midrule
1 & 2444 & 2168 & 276 & 27 & 1.25\%\\
2 & 2070 & 1593 & 477 & 23 & 1.44\%\\
3 & 1516 & 1061 & 455 & 19 & 1.79\%\\
4 & 1016 & 0 & 1016 & 0 & N/A\\
\rowcolor{rtpTableHighlight}
\multicolumn{1}{l}{Total} & 7046 & 4822 & 2224 & 69 & 1.43\%\\
\bottomrule\end{tabular}\end{table}

A selected-reuse exceedance exceeds either frozen fresh-relative tolerance after offline expansion; its denominator is selected reusable updates. The separate false-fresh rate conditions on at least one acceptable reusable candidate. Diagnostic intervals resample complete episodes within tasks, preserving all boundaries/candidates. Boundary-specific rates condition on survival and visited policy states. For default RTP under nominal conditions, false fresh is \DefaultFalseFresh/\DefaultAcceptable{} over states with at least one label-admissible reusable candidate. Structural exhaustion is excluded from this denominator.

\section{Observable Prediction and Loss Controls}
\label{app:probes}
\subsection{Common-Reference and Own-Action Probes}
The probes use two disjoint source pools, each with 20 eligible keys/task. In the common-reference pool, a fresh source controller runs under nominal, four-sample, or eight-sample hold conditions, generates a parent at boundary 12, and provides the physical state cloned at 16. Histories through 12 are common for the two hold durations, while feedback over $[12,16)$ may differ. Within each condition, candidates start from exactly the same clone. A fresh behavior continuation supplies common observed targets; candidates are not executed for this metric.

Own-action probes use a fresh parent generated at boundary 8, a four-sample hold, and a clone at 12. Retention, fresh, reconstruction-10, and Bridge-10 each produce one candidate. Each clone then executes its own action-conditioned continuation with updated facts and no further visual revision, stopping the diagnostic at the old window end 24. All predictions, including fresh, are compared on endpoints 16, 20, and 24. Fresh's endpoint 28 is excluded. The accepted prefixes are re-decoded with current facts at each intervening boundary.

Own-action metrics comprise contact-time absolute error, exact event-order accuracy, and latent error against each branch's own executed observations. Stratifying by whether the old order agrees with native's realized outcome is descriptive, because this stratum depends on one policy's outcome. Pairwise causal contrasts instead use the common pre-branch source keys.

MAE denotes mean absolute error; contact MAE is measured in native environment samples.
\begin{table}[H]\centering\small
\caption{Prediction against each candidate's own action-conditioned continuation, 320 complete windows per branch; contact mean absolute error (MAE) is in native samples.}\label{tab:own-action-prediction}
\begin{tabular}{lccc}\toprule
Candidate & Contact MAE & Order (\%) & Visual error\\\midrule
Retain & 3.02 & 59.69 & 0.1251\\
Fresh & 2.01 & 74.38 & 0.0893\\
Reconstruction-10 & 2.43 & 70.31 & 0.0955\\
Bridge-10 & 1.63 & 79.06 & 0.0691\\
\bottomrule\end{tabular}\end{table}

\subsection{Eligibility, Annotation and Losses}
Source selection sorts a separately reserved reset pool and takes the first 20 nonterminal sources/task reaching the required snapshot before candidate generation. Paired eligibility requires both hold conditions to reach that snapshot. This source eligibility applies only to diagnostics; primary endpoints retain every reset.

A task-success flag can be recorded while the diagnostic continues to its fixed window end if the benchmark permits continued safe state evolution. Invalid-state or safety termination stops that branch immediately. The source remains in the diagnostic pool with unavailable observations masked and attrition counted by branch. Paired latent comparisons use the common observed time intersection. Complete-window event/contact metrics use the common eligible subset, accompanied by its denominator and truncation counts; all metric targets are observed values.

Two annotators blind to method identify first visible target contact and task-script predicate order; a third adjudicates disagreements. The active target is fixed before cloning as the first unsatisfied interaction predicate's object, with ties broken by object index. Annotation preserves object identities and repeated events, with script-order ties for simultaneous transitions. Recorded environment event flags may aid adjudication of realized frames only. Missing contact in a fully observed twelve-sample window is assigned the first sample after the window (25 for the 12--24 example). Truncated windows have missing labels, not this sentinel.

Loss controls refit separate native corrections on identical data and schedules, changing only one coefficient in Equation~\ref{eq:bridge-loss} to zero. Evaluation uses Fixed bridge-10 without a selector, with current facts and the resulting prefix. Terminal success and common-reference visual error are separate endpoints: they test different targets, so their orderings need not agree.
To isolate learned correction from ordinary feedback retargeting, Fixed bridge-10 with zero residual uses the same native-checkpoint construction and uses current facts but sets the added velocity residual identically to zero. It shares source storage, depth, action decoding, and exhaustion rules with learned Fixed bridge-10. The two policies share reset/initial randomness and the checkpoint-storage protocol; later checkpoint tensors and observations may differ because both policies run their own closed-loop trajectories; this contrast is distinct from changing the checkpoint source or removing one training-loss term.
\begin{table}[H]\centering\small
\caption{Zero versus learned residual at fixed Fixed bridge-10. Source construction, depth, and action interface are matched; differences are learned minus zero.}\label{tab:bridge-learning}
\begin{tabular}{lccc}\toprule
Condition & Zero (\%) & Learned (\%) & Difference (pp)\\\midrule
Nominal & 41.62 & 47.38 & 5.75\\
Hold 4 & 41.62 & 47.38 & 5.75\\
Hold 8 & 40.75 & 47.12 & 6.38\\
\bottomrule\end{tabular}\end{table}

\begin{table}[H]\centering\small
\caption{Training-loss ablations at fixed Fixed bridge-10. Success and visual error have different targets; differences are full minus row.}\label{tab:loss-terms}
\begin{tabular}{lccc}\toprule
Training loss & Success (\%) & Visual error & Full--row (pp)\\\midrule
\rowcolor{rtpTableHighlight}
All terms & 47.38 & 0.0840 & 0.00\\
No observed-visual loss & 44.38 & 0.1206 & 3.00\\
No action loss & 44.88 & 0.0805 & 2.50\\
No fresh regularizer & 46.12 & 0.0961 & 1.25\\
\bottomrule\end{tabular}\end{table}

\section{Mismatch and Interface Sensitivity}
\label{app:stress}
\subsection{Holds and Source Comparisons}
Every primary reset has an intervention scheduled once at sample 8. The actuator holds the joint target from sample 7 for $d=4$ or 8 samples, preserves the currently commanded gripper coordinate, and resumes joint commands at $8+d$. Observation and proprioception updates continue throughout. The factual update receives the actual applied controls, while neither the timer nor duration is supplied as a feature. The endpoint includes early terminations and unexposed episodes, preserving the assigned intervention cohort.
\begin{table}[H]\centering\small
\caption{Complete-policy comparisons under actuator holds, 800 keys/condition. Differences are RTP minus Adaptive recon. Recon. denotes reconstruction from independent noise.}\label{tab:feedback-stress}
\begin{tabular}{cccccc}\toprule
Hold & Fresh & Fresh-20+C & Adaptive recon. & RTP & Difference (pp)\\\midrule
None & 40.38 & 43.50 & 45.62 & 48.63 & 3.00\\
4 & 40.38 & 43.50 & 45.62 & 48.63 & 3.00\\
8 & 39.62 & 43.12 & 44.75 & 48.25 & 3.50\\
\bottomrule\end{tabular}\end{table}

\begin{table}[H]\centering\small
\caption{Controller work under actuator holds. Noninitial calls use each policy's own denominator. Recon. denotes reconstruction from independent noise.}\label{tab:stress-cost}
\begin{tabular}{lcc}\toprule
Condition / policy & Mean steps & Mean call (s)\\\midrule
4 / RTP & 8.28 & 1.060\\
4 / Adaptive recon. & 8.24 & 1.065\\
8 / RTP & 8.88 & 1.071\\
8 / Adaptive recon. & 8.95 & 1.078\\
4 / Fixed bridge-10 & 12.33 & 1.115\\
4 / Fixed recon.-10 & 12.32 & 1.117\\
8 / Fixed bridge-10 & 12.33 & 1.115\\
8 / Fixed recon.-10 & 12.32 & 1.117\\
\bottomrule\end{tabular}\end{table}

\begin{table}[H]\centering\small
\caption{Original-root reconstruction and RTP. Differences are RTP minus Adaptive root recon.; step counts include all refreshes. Root recon. denotes reconstruction with the original root noise.}\label{tab:adaptive-root}
\begin{tabular}{lccccc}\toprule
Condition & Adaptive root recon. (\%) & RTP (\%) & \shortstack{Root recon.\\steps} & RTP steps & Difference (pp)\\\midrule
Nominal & 46.12 & 48.63 & 8.07 & 7.78 & 2.50\\
Hold 4 & 46.12 & 48.63 & 8.65 & 8.28 & 2.50\\
Hold 8 & 45.50 & 48.25 & 9.27 & 8.88 & 2.75\\
\bottomrule\end{tabular}\end{table}

Ten-interval source controls share operation depth and exhaustion rules, but their achieved refresh fractions can differ when episode lengths differ. Cost accounting uses each policy's counts and times; equal depth does not imply equal whole-policy cost. All models and thresholds remain frozen.

\subsection{Activation Delay}
At each noninitial boundary $b$, the controller captures facts through $b$ once and generates a block $(u_b,\ldots,u_{b+3})$ from this capture. During $[b,b+D)$, dynamics evolve under the preceding target; these later observations are inaccessible to the in-flight call. At arrival, the executor discards the first $D$ expired action slots and executes the remaining commands at their original timestamps. At $b+4$, the factual update incorporates all applied controls/observations, and the original prediction is compared at that same endpoint.

The delay settings are $D=0,1,2$, with nominal parameters frozen. Termination ends both waiting and execution. A deterministic buffered implementation may calculate the output first provided it cannot read intervening observations. This imposes the same capture delay across policies independently of their measured GPU time. $D\ge4$ misses the entire action block and is outside this driver.
\begin{table}[H]\centering\small
\caption{Activation-delay comparisons on 800 reset keys per delay. Differences are RTP minus Adaptive recon. Recon. denotes reconstruction from independent noise.}\label{tab:activation-delay}
\begin{tabular}{cccccc}\toprule
Delay $D$ & Fresh-20+C & Fresh-10+C & Adaptive recon. & RTP & Difference (pp)\\\midrule
0 & 43.50 & 41.75 & 45.62 & 48.63 & 3.00\\
1 & 40.88 & 40.25 & 42.38 & 46.62 & 4.25\\
2 & 37.50 & 37.62 & 38.62 & 43.38 & 4.75\\
\bottomrule\end{tabular}\end{table}

\subsection{Spatial Shift and Temporal Configuration}
The fixed task subset is T01,T03,T05,T06,T09,T10,T13,T16 (400 keys). At boundary 8 after capture and before action, the intervention translates the free target 2 cm horizontally, with sign from reset-key parity. Direction and clearance rules are fixed on development states. A constrained or colliding target receives no displacement, while its reset remains in the endpoint. Exposure counts are policy-specific, and subset membership is independent of post-intervention success.
\begin{table}[H]\centering\small
\caption{Spatial-shift comparisons on 400 reset keys. Complete-policy differences use RTP; fixed-depth differences use Fixed bridge-10. Recon. denotes reconstruction from independent noise.}\label{tab:geometry-shift}
\begin{tabular}{lccc}\toprule
Policy & Success (\%) & Mean steps & Matched diff. (pp)\\\midrule
Fresh-20+C & 38.75 & 20.00 & 6.25\\
Adaptive recon. & 41.25 & 8.94 & 3.75\\
\rowcolor{rtpTableHighlight}
RTP & 45.00 & 9.03 & 0.00\\
Fixed recon.-10 & 39.50 & 12.31 & 3.75\\
Fixed bridge-10 & 43.25 & 12.32 & 0.00\\
\bottomrule\end{tabular}\end{table}

The $H=6,J=4$ setting uses a base, bridge, and estimator refitted with six complete groups; the lifecycle admits five reusable updates. The $H=4,J=8$ setting uses four evenly spaced observations per action block and decodes eight native actions/group. This changes environment-sample spacing and action-head alignment, so every comparator is adapted under the new convention. History endpoints are scaled by the new $J$, budgets remain in groups, and training manifests and budgets are matched within each setting.
Each interface setting uses its own adapted base, bridge, and estimator.
\begin{table}[H]\centering\small
\caption{Horizon and sample-spacing comparisons on 800 reset keys per interface. Entries are success percentages. Recon. denotes reconstruction from independent noise.}\label{tab:window-spacing}
\begin{tabular}{lccccc}\toprule
Setting & Fresh & Fresh-20+C & Adaptive recon. & RTP & RTP--Adaptive recon. (pp)\\\midrule
$H=6,J=4$ & 42.12 & 45.00 & 47.25 & 50.62 & 3.38\\
$H=4,J=8$ & 37.62 & 40.12 & 41.38 & 44.12 & 2.75\\
\bottomrule\end{tabular}\end{table}

\section{Controller and Resource Accounting}
\label{app:resources}\label{app:source-coupling}
The hardware configuration is one NVIDIA H200 141-GB GPU with batch-one bfloat16 inference. Timing metadata comprise framework, CUDA and driver versions, checkpoint identifiers, and hardware identifiers. Timed calls are bracketed by device synchronization and follow 100 discarded warmups. Full-call timing covers factual selection/encoding, positions/masks, scoring, selected visual generation and action decoding; image transport, rendering and environment stepping are separate timing components. Primary timing pauses dynamics; imposed delays follow their native-sample schedule. The timing results refer to RoboMME.

Mean time uses each policy's noninitial call count as its denominator; initialization has a separate entry and contributes to episode time. P50/P95 are the 50th/95th percentiles of call duration. Cost accounting ends at termination, and paired episode-cost intervals use bootstrap resampling by reset key.

Let $f_k$ be the measured fresh fraction of fixed-$k$ calls. Let $n_0,n_5,n_{10},n_T$ count retention, Bridge-5, Bridge-10, and fresh calls among $B$ noninitial decisions. The analytic cost identities are
\begin{equation}
n_0+n_5+n_{10}+n_T=B,\qquad
\overline N_{\rm RTP}=\frac{5n_5+10n_{10}+20n_T}{B},\qquad
\overline N_k=k+(20-k)f_k.
\label{eq:cost}
\end{equation}
Actual backbone forward counts include guidance multiplicity and are distinct from solver-interval counts. For a selected branch, only one next-action decode is executed.
\begin{table}[H]\centering\small
\caption{Controller-call durations. Policy rows contain noninitial calls; the last row contains the 800 shared fresh initializations. Quantiles use nearest rank.}\label{tab:latency-details}
\begin{tabular}{lccc}\toprule
Policy & Mean (s) & P50 (s) & P95 (s)\\\midrule
Fresh & 1.248 & 1.248 & 1.304\\
Fixed bridge-5 & 1.057 & 1.007 & 1.281\\
Fixed bridge-10 & 1.115 & 1.088 & 1.268\\
\rowcolor{rtpTableHighlight}
RTP & 1.051 & 1.010 & 1.294\\
Shared initialization & 1.277 & 1.278 & 1.335\\
\bottomrule\end{tabular}\end{table}

\begin{table}[H]\centering\small
\caption{Mode counts and episode cost on 800 episodes. Episode time includes initialization and all noninitial calls.}\label{tab:cost-details}
\begin{tabular}{lcccccc}\toprule
Policy & Updates & Retain & Bridge-5 & Bridge-10 & Fresh & s/episode\\\midrule
Fresh & 17403 & 0 & 0 & 0 & 17403 & 28.430\\
Fixed bridge-5 & 17882 & 0 & 13718 & 0 & 4164 & 24.902\\
Fixed bridge-10 & 17947 & 0 & 0 & 13769 & 4178 & 26.293\\
\rowcolor{rtpTableHighlight}
RTP & 17922 & 6574 & 4407 & 2139 & 4802 & 24.817\\
\bottomrule\end{tabular}\end{table}

First-order Euler stores two checkpoints and the complete clean endpoint: $3\cdot2HPC$ bfloat16 bytes. For $H=4,C=48,P=512/480$, this is \RoboStateBytes{}/\RmbStateBytes{} bytes (\RoboStateMiB{}/\RmbStateMiB{} MiB). Bridge-5 stores one checkpoint plus endpoint: \RoboBridgeFiveBytes{}/\RmbBridgeFiveBytes{} bytes. Adaptive root-noise reconstruction regenerates root noise from its identifier. Metadata, weights, factual history, encoder cache and temporary activations are counted separately.

Including biases, parameter counts for DiffEncode, residual conditioning, the velocity correction, and the discrepancy estimator are
\begin{equation}
\begin{aligned}
P_{\rm Diff}&=DD_\Delta+D^2+2D,&
P_{\rm cond}&=D(D+2)+D,\\
P_{\rm res}&=(D_b+D)D+D+DC_p+C_p,&
P_{\rm sel}&=2D_S^2+6D_S+4.
\end{aligned}
\label{eq:parameters}
\end{equation}
The residual has \ResidualParameters{} parameters. Bridge+DiffEncode totals are \RoboBridgeParameters{}/\RmbBridgeParameters{}; the discrepancy estimator has \SelectorParameters{}; combined totals are \RoboTotalParameters{}/\RmbTotalParameters{}. Fixed pooling and unpatchification add none. Removing age/length changes $D_S$ and the corresponding parameter count.

\end{document}